\documentclass[10pt,twocolumn,letterpaper]{article}

\usepackage[pagenumbers]{cvpr} 

\usepackage{times}
\usepackage{epsfig}
\usepackage{graphicx}
\usepackage{amsmath}
\usepackage{amssymb}
\usepackage{bbm}
\usepackage{multirow}
\usepackage{adjustbox}
\usepackage{booktabs}
\usepackage{fontenc}
\usepackage{enumitem}
\usepackage{dirtytalk}
\usepackage[dvipsnames]{xcolor}
\usepackage{microtype}
\usepackage{bm}
\usepackage{float} 
\usepackage[sectionbib]{bibunits}
\usepackage{dblfloatfix}

\usepackage[ruled,linesnumbered]{algorithm2e}
 \newcommand{\Hquad}{\hspace{0.5em}}

\usepackage[dvipsnames]{xcolor}

\definecolor{cvprblue}{rgb}{0.21,0.49,0.74}
\usepackage[pagebackref,breaklinks,colorlinks,citecolor=cvprblue]{hyperref}

\def\paperID{26} 
\def\confName{CVPR}
\def\confYear{2024}

\title{Image Outlier Detection Without Training using RANSAC}

\author{Chen-Han Tsai\\
HTC DeepQ\\
{\tt\small maxwell\_tsai@htc.com}
\and
Yu-Shao Peng\\
HTC DeepQ\\
{\tt\small ys\_peng@htc.com}
}

\begin{document}
\maketitle
\begin{abstract}
Image outlier detection (OD) is an essential tool to ensure the quality of images used in computer vision tasks. 
Existing algorithms often involve training a model to represent the inlier distribution, and outliers are determined by some deviation measure.
Although existing methods proved effective when trained on strictly inlier samples, their performance remains questionable when undesired outliers are included during training.
As a result of this limitation, it is necessary to carefully examine the data when developing OD models for new domains.
In this work, we present a novel image OD algorithm called RANSAC-NN that eliminates the need of data examination and model training altogether.
Unlike existing approaches, RANSAC-NN can be directly applied on datasets containing outliers by sampling and comparing subsets of the data.
Our algorithm maintains favorable performance compared to existing methods on a range of benchmarks. 
Furthermore, we show that RANSAC-NN can enhance the robustness of existing methods by incorporating our algorithm as part of the data preparation process.

\end{abstract}

\section{Introduction}
\label{sec:intro}

Outlier detection (OD) refers to the task of identifying abnormal data that deviates from a given target distribution. 
Abnormal data are known as \textit{outliers}, and \textit{inliers} are data that belong within the target distribution. In general, the outlier distributions are not known in advance, and the goal of OD methods is to separate inliers from outliers in a \textit{one-class classification} fashion \cite{one_class_classification, oc-svm, 2004_svdd, 2019_deep_feature_for_oneclass_cls, a_surve_of_one_class_khan}.

Image OD is an extension of general OD for the field of computer vision  \cite{od_survey, generalize_ood_survey, ImageOD-survey}. The objective is to identify complete images or patches of images that are visually distinct. Low-level pixel regions that protrude from neighboring patches are considered patch-level abnormalities. Abnormality at the image level is an image that looks different from a regular image as a whole. For the scope of this paper, we limit our discussion towards OD at the image level.

The variety of image OD methods ranges from traditional tabular techniques to recent deep learning-based approaches. Traditional techniques often assume the data to consist of low-dimensional tabular features \cite{isolation_forest, algo-inne, local_outlier_factor, loda}.  However, these methods have not been designed to handle the high-dimensional pixel array in natural images. Deep learning introduced an automated method of extracting meaningful, resolution-independent features directly from images. This allowed the use of feature embeddings extracted by pre-trained neural networks to replace traditional features used by tabular methods \cite{knn-distance-algo, iForest_on_NN, 2019_deep_feature_for_oneclass_cls}. Neural network methods that learn unique features of the inlier distribution have also been proposed \cite{2018_deepsvdd, anogan, 2015_vae_anomaly_det, geom, goad}.

A big challenge among existing image OD methods is the dependence on a \textbf{clean inlier set} for model training. To build a good representation of the inlier distribution, the training data must contain enough inlier samples while avoiding outlier \textit{contamination}. Outlier presence causes actual outliers to be incorrectly treated as inliers during training, which leads to deteriorated OD performance during test time. Manual inspection is often necessary to ensure data quality, but this can be a time-consuming and expensive process.  

In this paper, we present RANSAC-NN, a novel image OD algorithm that addresses outlier contamination in image datasets. Inspired by RANSAC \cite{ref_ransac}, which was intended for the fundamental matrix estimation, our algorithm applies a two-stage iterative nearest neighbor sub-sampling for outlier prediction. Specifically, Inlier Score Prediction (ISP) in the first stage generates a set of \textit{inlier score} estimations, which are used by the second stage Threshold Sampling (TS) to predict the \textit{outlier score} of each sample. Outliers in the contaminated set can be removed by filtering out samples with outlier scores exceeding a certain threshold. Unlike previous approaches, whose objective is to develop a task-specific OD model for inference purposes, the goal of RANSAC-NN is to quantify outlier samples in a given contaminated set based on its image distribution alone. As a result, our algorithm does not require training or data preparation and can be applied straight out of the box.

We evaluate RANSAC-NN in comparison to existing methods on a range of OD tasks. The first observation is that almost all existing methods achieve similar performance when trained on a clean inlier set. However, when the training set is contaminated by outliers, a noticeable drop in performance occurs depending on the amount of outlier contamination. We then attempt to filter the outlier samples by applying RANSAC-NN on the contaminated sets. Models of existing OD algorithms were trained using the filtered datasets, and the results show significant improvements in all the algorithms. Additional experiments regarding the RANSAC-NN setup and hyper-parameter settings have also been explored in our evaluations.

In summary, we make the following contributions:
\begin{itemize}

    \item We demonstrate that the majority of image OD algorithms can achieve favorable and robust performance when trained properly using a clean inlier set. 
    
    \item We emphasize the importance of training on a clean inlier set by examining the performance drop of existing methods when trained on a contaminated dataset. 

    \item We present a novel image OD algorithm called RANSAC-NN that delivers promising performance in comparison to existing methods on a number of benchmarks. Our algorithm does not require data preparation or model training, and it is effective on datasets contaminated by outliers.

    \item We demonstrate an improved robustness against outlier contamination in all existing OD algorithms by employing RANSAC-NN during data preparation. 
    
\end{itemize} 

The remaining of this paper is organized as follows. In Section \ref{sec:related_works}, we provide an overview of some well-known OD methods. In Section \ref{sec:methods}, we present the proposed RANSAC-NN algorithm. In Section \ref{sec:experiments}, we detail the experimental results. Finally, we conclude the paper in Section \ref{sec:conclusions}.

\section{Related Works}
\label{sec:related_works}

Image OD methods can be categorized by their detection mechanism into three main types: density estimation, image reconstruction, and self-supervised classification. We provide a brief introduction of each in the following.

\subsection{Density Estimation}
Density estimation relies on a set of inlier images to build a representation of the inlier distribution. During test time, images are compared to the representation to determine their outlier scores. While some methods have suggested to explicitly model the inlier distribution \cite{book-prml-bishop, a_surve_of_one_class_khan, 2018_dae_gmm}, others have explored non-parametric methods. 

One strategy is to combine tabular non-parametric methods with neural networks to address outlier images. In one study \cite{iForest_on_NN}, image embeddings extracted by a neural network were fed into the Isolation Forest \cite{isolation_forest} and Local Outlier Factor (LOF) \cite{local_outlier_factor} algorithms for outlier score prediction. Other studies \cite{knn-distance-algo, 2019_deep_feature_for_oneclass_cls} have considered the use of nearest neighbors between image embeddings as an outlier score measure. 

Deep learning-based density estimation methods rely on optimizing a neural network to learn characteristics specific to inlier images. DeepSVDD \cite{2018_deepsvdd} is a method inspired by \cite{oc-svm, 2004_svdd} that learns a mapping from the inlier image distribution to a minimum volume hypersphere in latent space. At test time, images that are mapped to points far from the center of the hypersphere are considered likely outliers. OC-CNN \cite{2018_oneclass_cnn} suggested a similar approach, but it leveraged a pre-trained convolutional neural network that can be trained in an end-to-end fashion. Adversarial techniques \cite{sogann} that combine synthetic outlier images with real inlier images to train a discriminator network have also been presented. Given the wide variety of density estimation methods, we refer to \cite{ImageOD-survey, od_survey} for a complete overview.

\subsection{Image Reconstruction}
Image reconstruction refers to a class of methods that measures the outlier score of an image by its reconstruction \cite{Outlier_Analysis_aggarwal}. Auto-encoders and generative adversarial networks (GAN's) are two common options given their ability to generate realistic images using latent representations \cite{2014_anomaly_detection, 2014_gans}. In \cite{2015_vae_anomaly_det, 2015_autoencoder_reconstruction_loss}, auto-encoders were trained on a set of inlier images, and the outlier scores of test images were measured by comparing their reconstruction loss with those of known inlier images. This method was further improved in \cite{2019_as_ae, 2019_mem_ae} by including additional latent heuristics as part of the comparison. Other works \cite{2020_attr, 2020_puzzle_ae} have suggested the addition of image augmentations to increase the distinction between latent representations.  

GAN-based reconstruction relies on the generative ability of a generator network to mimic the inlier image space. One approach \cite{deecke2018anomaly} is to search for a latent representation of a test image from a trained generator. At test time, images that fail to find a corresponding latent representation are considered outliers. Another method \cite{anogan} attempts to generate a replication of a given image using a trained generator. The image is considered an outlier if the generator fails to replicate a similar reconstruction.

\subsection{Self-Supervised Classification}
Self-supervised classification is a type of method that follows the same principles from self-supervised learning \cite{ssl_cookbook}. In self-supervised learning, a model is optimized on one or more auxiliary tasks using transformed data from an unlabeled dataset. The purpose is to learn meaningful representations that can potentially benefit downstream tasks by training on the auxiliary tasks. 

GEOM \cite{geom} proposed an auxiliary objective to train a classification model in identifying geometric transforms on unlabeled inlier images. During inference, the same transformations are applied to the test image, and by comparing the model's output with those seen during training, the outlier score can be determined. Additional transforms were introduced by \cite{2019_using_ssl_can_improve_robustness}, which have demonstrated improved OD performance. GOAD \cite{goad} builds upon GEOM by generalizing the types of transforms to include non-image data. Additional methods \cite{2020_ssl_ae, 2020_csi_ssl_ae, 2021_ssd_od, 2021_neural_transformation} have also shown promising performance using self-supervised classification methods.

\section{Methods}
\label{sec:methods}

For a dataset $D$ containing $n$ images, a set of image embeddings $F = \{ f_i \}_{i=1}^n$ are extracted by a feature extractor. Both inlier and outlier images are present in $D$. \textbf{The assumption is that inliers to constitute the majority in $D$}. The goal of RANSAC-NN is to assign outlier scores $\sigma$ approaching $1$ to the outlier images and $\sigma$ approaching $0$ to the inlier images. 

\subsection{RANSAC-NN}
\label{methods-ransacnn_algo}
The RANSAC-NN algorithm consists of two main components: Inlier Score Prediction (ISP) and Threshold Sampling (TS). 
ISP identifies highly probable inliers in the dataset when the distribution is initially unknown. It begins by sampling the embedding set $F = \{ f_i \}_{i=1}^n$ for a number of iterations. The sampled results are then aggregated, and ISP assigns a value $\eta_i \in [0, 1 ]$ to each embedding $f_i$ that indicates the likelihood of the $i$-th image to be an inlier. This value $\eta_i$ is referred to as the \textit{inlier score} of the $i$-th image. 

The inlier scores $\boldsymbol{\eta} = \{ \eta_i \}_{i=1}^n$ predicted by ISP estimates the inlier distribution of $F$. TS uses the inlier scores $\boldsymbol{\eta}$, and it samples $F$ repeatedly for a second time. This time, however, TS selectively samples highly-probable inliers in $F$ by only choosing candidates with inlier scores $\eta$ above a certain threshold $\tau$. This process is repeated for a range of threshold values, and the results are aggregated to compute the final outlier scores $\sigma$ of each image. This sampling strategy allows TS to predict outlier scores $\boldsymbol{\sigma}$ with greater precision in comparison to ISP\footnote{A possible outlier score measure could be the inverse of the inlier scores (i.e., $1-\eta$). This measure only requires the ISP stage, but it offers less precision compared to the outlier scores $\sigma$ predicted by TS.}. In Section \ref{exp:ablation_studies} of our experiments, we show the importance of the TS stage in enhancing the predicted outlier scores. 

\paragraph{Inlier Score Prediction (ISP)}
\label{isp_stage}

The objective of ISP is to predict a set of inlier scores $\boldsymbol{\eta} = \{ \eta_i \}_{i=1}^n$ for a given set of image embeddings $F = \{ f_i \}_{i=1}^n$. Since the inlier distribution is initially unknown, ISP samples $F$ uniformly in an attempt to obtain a set of embeddings $\Tilde{I}$ consisting of all inliers, which we refer to as a \textit{clean} set. 
In total, ISP samples $F$ for $s$ iterations, each time taking $m$ embeddings without repetition. This set of $m$ sampled embeddings, denoted as $\Tilde{I}$, are assumed to all be inliers, and they are compared against the remaining embeddings in $F$. Comparison is made by taking the cosine similarities for each $f_i \in F$ to its nearest-neighbor (NN) embedding in the sample set $\Tilde{I}$. 
A similarity score $\alpha_i \in [-1, 1]$ stores this value for every $f_i \in F$.
An embedding $f_i$ with an $\alpha_i$ approaching $1$ indicates that $f_i$ is very similar to one of the sampled $\Tilde{f} \in \Tilde{I}$. Moreover, if $\Tilde{I}$ is a clean set, inliers in $F$ would all have a high value of $\alpha_i$. 

In reality, however, the sample set $\Tilde{I}$ may contain outliers. Thus, by repeatedly sampling $F$, we can increase the odds of obtaining one clean set $\Tilde{I}$ during one of the $s$ sampling iterations. For each $f_i \in F$, the similarity scores $\alpha_i$ obtained in each iteration are aggregated by taking the minimum $\alpha_i$ obtained across all $s$ iterations. This minimum $\alpha_i$ of each $f_i \in F$ essentially measures the worst-performing similarity score from all $s$ iterations. As long as an inlier is present in every sampled $\Tilde{I}$, a lower bound of the minimum $\alpha_i$ for $f_i \in F_\text{in}$ exists. On the contrary, if an outlier $f_i \in F_\text{out}$ ever attains a low $\alpha_i$ during one of the $s$ iterations, which would happen if $\Tilde{I}$ is a clean set, the minimum $\alpha_i$ would account for this drop in similarity score. This minimum $\alpha_i$ is referred to as the inlier score $\eta_i$ for image $i$. The details of ISP are provided in Algorithm \ref{alg:isp}. 


\SetKwProg{Init}{init}{}{}
\SetKwFor{RepTimes}{repeat}{times}{end}
\SetKwInOut{Input}{Input}\SetKwInOut{Output}{Output}


\begin{algorithm}[t]

\caption{Inlier Score Prediction (ISP)}
\label{alg:isp}

\Input{Embeddings $F = \{f_i\}_{i=1}^n, \Hquad f_i \in \mathbb{R}^d$\\
Sampling iterations $s$ \\
Sample size $m$\\
}
\Output{Inlier scores $\boldsymbol{\eta} = \{\eta_i\}_{i=1}^n, \Hquad \eta_i \in [0, 1]$}

\Init{}{
    $\boldsymbol{\eta} = \{ \eta_i \}_{i=1}^n$ where $\eta_i = 1$ \\
}

\RepTimes{$s$}{
    $\Tilde{I} \gets $ take $m$ samples from $F$\\
    \For{$i \gets 1$ to $n$}{
        $\alpha_i = \max{( \{ \cos{(f_i, \Tilde{f})} \mid \Tilde{f} \in \Tilde{I}\} )}$ \\
        $\eta_i \gets \min{( \eta_i,  \alpha_i)}$
    }
}

\Return{$\boldsymbol{\eta}$}
\end{algorithm}

\paragraph{Threshold Sampling}
\label{threshold_sampling}

TS leverages $\boldsymbol{\eta}$ from ISP to predict the final set of outlier scores $\boldsymbol{\sigma} = \{ \sigma_i \}_{i=1}^n$ for each embedding in $F = \{ f_i \}_{i=1}^n$. Unlike ISP, where the inlier distribution is initially unknown, TS has the advantage of using the inlier scores $\boldsymbol{ \eta }$ as a prior. The idea behind TS is to compare each $f_i \in F$ starting from a loosely assembled set of inlier candidates to a gradually refined set. This gradual transition allows for inliers that might have a relatively lower $\eta_i$ to still be included in the sample set $\Tilde{I}$ during early sampling iterations. Inliers with larger $\eta_i$, or embeddings that resemble most of the inliers, would be sampled frequently. 

In each sampling iteration, a threshold $\tau \in [0, 1]$ filters out embeddings in $F$ that have an inlier score $\eta_i \leq \tau$. This leaves a set of eligible embeddings $\Omega = \{f_i \mid \eta_i > \tau \} \subseteq F$ available for sampling. Similar to ISP, the algorithm takes at most $m$ samples uniformly from $\Omega$ and assembles a set $\Tilde{I}$. In contrast to ISP, $\Tilde{I}$ in TS would contain a higher concentration of actual inliers since $\Omega$ has already filtered out embeddings with low $\eta_i$. 
Each embedding $f_i \in F$ is then compared against the sample set $\Tilde{I}$, and the similarity score $\alpha_i$ is computed for each $f_i \in F$. 
For each embedding $i$, a running mean $\sigma_i$ keeps track of the average number of times the embedding $f_i$ obtains an $\alpha_i < \tau$. 

This sampling and comparison process is repeated for every increase in threshold $\tau$. The final $\sigma_i \in [0, 1]$ is the outlier score predicted by TS, which equals the average number of times each $f_i \in F$ fails to find a close-enough NN during TS. An $f_i \in F$ with $\sigma_i$ approaching $1$ implies that the embedding failed during the majority of TS sampling iterations, which is a common characteristic of outliers. Details of TS are listed in Algorithm \ref{alg:threshold_sampling}.

\label{algorithm-threshold_sampling}

\SetKwProg{Init}{init}{}{}
\SetKwFor{RepTimes}{repeat}{times}{end}
\SetKwInOut{Input}{Input}\SetKwInOut{Output}{Output}

\begin{algorithm}[t]

\caption{Threshold Sampling (TS)}
\label{alg:threshold_sampling}

\Input{Embeddings $F = \{f_i\}_{i=1}^n, \Hquad f_i \in \mathbb{R}^d$\\
Inlier scores $\boldsymbol{\eta} = \{ \eta_i \}_{i=1}^n, \Hquad \eta_i \in [0, 1]$ \\
Threshold iterations $t$\\
Sample size $m$\\
}
\Output{Outlier scores $\boldsymbol{\sigma} = \{\sigma_i\}_{i=1}^n, \Hquad \sigma_i \in [0, 1]$}

\Init{}{
    $\boldsymbol{\sigma} = \{ \sigma_i \}_{i=1}^n$ where $\sigma_i = 0$ \\
}

\For{$k \gets 1$ to $t$}{
    $\tau \gets (k-1)/t$ \\
    $\Omega \gets \{ f_j \mid \eta_j > \tau, f_j \in F \} \quad$ \\
    \eIf{$ | \Omega | > 0 $}{
        $\Tilde{I} \gets$ take $\min{(| \Omega |, m)}$ samples from $\Omega$ \\
        \For{$i \gets 1$ to $n$}{
            $\alpha_i \gets \max{(  \{  \cos{(f_i , \Tilde{f} )} \mid \Tilde{f} \in \Tilde{I}\}  )}$ \\
            $\sigma_i \gets (\mathbbm{1}{ [ \alpha_i < \tau ] } + (k-1) \cdot \sigma_i)/k$ \\
        }
    }{
        \textbf{break}
    }
}

\Return{$\boldsymbol{\sigma}$}
\end{algorithm}

\subsection{Properties Analysis}
\label{methods:sample_size_iteration_properties}
In this section, we explore the properties of RANSAC-NN in detail. Assumptions regarding the inlier and outlier embeddings are presented, and we analyze the different possibilities of obtaining a sample set $\Tilde{I}$ during ISP. We then analyze the influence of the sample size $m$ and sampling iterations $s$ on RANSAC-NN. The sample size $m$ determines how large of a set $\Tilde{I}$ to sample for, and the sampling iteration $s$ controls the probability of obtaining a clean inlier set. Since $m$ and $s$ mainly influence the inlier scores $\boldsymbol{ \eta }$, we focus on ISP for most of this section.

For the purpose of analysis, we assume that $F = F_{\text{in}} \cup F_{\text{out}}$ and $F_\text{in} \cap F_\text{out} = \varnothing$ where $F_{\text{in}}$ and $F_{\text{out}}$ are the inlier and outlier subsets of $F$. Assume further that all embeddings $f_i \in F$ satisfy the following bounds:
\begin{equation}
    \label{eq-assumption1}
    \cos{(f_1, f_2 )} > g \quad \forall \Hquad f_1, f_2 \in F_{\text{in}}
\end{equation}
\begin{equation}
    \label{eq-assumption2}
    \cos{(f_1, f_3 )} < h \quad \forall \Hquad f_1 \in F_{\text{in}}, f_3 \in F_{\text{out}}
\end{equation}
where $-1 \leq h \ll g \leq 1$. The bound in Equation \ref{eq-assumption1} presumes that all inlier embeddings in $F_\text{in}$ have a pair-wise cosine similarity greater than $g$. Equation \ref{eq-assumption2} entails that the cosine similarity of an inlier embedding and an outlier embedding does not exceed the value of $h$. To account for similar outliers, we do not assume any bounds on the pair-wise similarity between outlier embeddings. 

\paragraph{Influence of sample sets on ISP}

During ISP, sets of embeddings $\Tilde{I}$ are randomly sampled from $F$. We consider the different cases of $\Tilde{I}$, and we examine their influence on the predicted inlier scores during ISP.

\subparagraph{(1) Samples are all inliers}
In the case where $\Tilde{I}$ is a clean set, the similarity score $\alpha_i$ effectively separates inliers from outliers by a large margin because $\alpha_i > g \Hquad \forall \Hquad f_i \in F_\text{in}$ and $\alpha_i < h \Hquad \forall \Hquad f_i \in F_\text{out}$ (Line $6$, Algorithm \ref{alg:isp}). Furthermore, if an inlier is present in every sampled $\Tilde{I}$ throughout all $s$ iterations during ISP, the inlier scores $\eta_i \Hquad \forall \Hquad f_i \in F_\text{in}$ would be lower-bounded by $g$, and $\eta_i \Hquad \forall \Hquad f_i \in F_\text{out}$ would be upper-bounded by $h$. This is because ISP takes the minimum $\alpha_i$ across all $s$ iterations as $\eta_i$ (Line $7$, Algorithm \ref{alg:isp}). Thus, if a sampled $\Tilde{I}$ is ever a clean set in one of the $s$ iterations during ISP, inliers and outliers can be distinguishable from the predicted inlier scores $\boldsymbol{\eta}$.

\subparagraph{(2) Samples contain both inliers and outliers}
Most of the sampled $\Tilde{I}$ during ISP fall under this case. When $\Tilde{I}$ contains both inlier and outlier samples, the similarity score $\alpha_i \Hquad \forall \Hquad f_i \in F_\text{in}$ are lower bounded by $g$ (Line $6$, Algorithm \ref{alg:isp}). The outliers $f_i \in F_\text{out}$, however, are not necessarily upper bounded by $h$ due to the presence of outlier samples in $\Tilde{I}$. But this is fine because the inlier score $\eta_i$ predicted by ISP is the minimum $\alpha_i$ from all $s$ sampling iterations. As long as ISP samples a clean set during one of the $s$ iterations (Case 1), the inlier scores $\eta_i$ of the outliers $f_i \in F_\text{out}$ will still be upper-bounded by $h$.

\subparagraph{(3) Samples are all outliers}
In the case where $\Tilde{I}$ contains only outliers, ISP would fail to produce meaningful inlier scores $\boldsymbol{\eta}$. The reason is that the similarity score $\alpha_i$ for all the inliers $f_i \in F_\text{in}$ would be upper bounded by $h$ when compared against a sample set $\Tilde{I}$ containing all outliers. The outliers $f_i \in F_\text{out}$, on the other hand, are not necessarily upper-bounded by $h$. The predicted inlier scores $\eta_i \Hquad \forall \Hquad f \in F_\text{in}$ by ISP, which is the minimum $\alpha_i$ across all $s$ sampling iterations, would thus be upper-bounded by $h$. This implies that the inlier scores $\boldsymbol{\eta}$ have failed to produce a meaningful distinction between inliers and outliers. Furthermore, the inliers $f_i \in F_\text{in}$ would be filtered out when the threshold $\tau$ exceeds $h$ during TS, leading to deteriorating outlier prediction performance. The situation where $\Tilde{I}$ contains all outliers should be avoided at all costs during ISP sampling.

\paragraph{Sample Size}
Now that all possible cases of $\Tilde{I}$ have been explored, we explain how setting the sample size $m$ can avoid sampling an $\Tilde{I}$ containing all outliers (Case 3) and improve the odds to obtaining a clean set (Case 1). We denote the probability to sample a clean set as $p_\text{clean}$ and the probability to sample a set containing all outliers as $p_\text{out}$. Suppose there exists $l$ outliers in a dataset of size $n$. During each sampling iteration, the probability of sampling a clean set is given by:
\begin{equation}
    \label{eq-pclean}
    p_\text{clean} = \frac{{n-l \choose m}}{{n \choose m}} \approx \left( 1-\frac{l}{n} \right)^m
\end{equation}
and the probability of sampling $\Tilde{I}$ with all outliers is given by:
\begin{equation}
    \label{eq-outliers}
    p_\text{out} = \frac{{l \choose m}{n-l \choose n-m}}{{n \choose m}} \approx \left( \frac{l}{n} \right)^m
\end{equation}

From Equations \ref{eq-pclean} and \ref{eq-outliers}, we notice that when $l < n/2$, $p_\text{out} < p_\text{clean} \Hquad \forall \Hquad m > 0$. This implies that in the case where outliers constitute the minority, the probability of sampling a clean set is always greater than that of sampling a set with all outliers. For large sample sizes $m$, $p_\text{out}$ converges to $0$ at a fast rate, which implies a reduced chance of outliers populating $\Tilde{I}$. $p_\text{clean}$ also approaches $0$, but at a slower rate than $p_\text{out}$, implying a reduced probability of sampling a clean set.

On the contrary, if $m$ is assigned an extremely small value, then $p_\text{clean}$ would greatly increase. This indicates an increased probability of obtaining a \text{clean} set. But doing so would also increase $p_\text{out}$. Since sampling an $\Tilde{I}$ containing all outliers is detrimental to ISP, we avoid setting extremely small values of $m$.

\paragraph{Sampling Iteration}
For a given sample size $m$, the probability to sample a clean set $p_\text{clean}$ is fixed for each iteration. Then for a certain confidence $c$ (e.g. $c = 0.95$), the minimum number of sampling iterations $s_\text{min}$ such that at least one clean set has been sampled is given by:

\begin{equation}
    \label{eq-sample_iter}
    s_\text{min} \geq \lceil \log_{1-p_\text{clean}}(1 - c) \rceil
\end{equation}

Thus, as long as $m$ is set at a reasonable sample size, increasing the number of sampling iterations $s$ only improves the odds of sampling a clean set. In Section \ref{exp:hyperparameters} of the experiments, we demonstrate the above principles on actual image datasets.

\paragraph{Threshold Iterations}
Threshold iterations $t$ in TSP controls the resolution for which samples are filtered. A high value of $t$ (e.g., $1000$) allows for finer filtering compared to a low value of $t$ (e.g., $10$). Depending on the density of the inlier score distribution predicted by ISP, it is suggested to increase $t$ until the predicted outlier scores $\boldsymbol{\sigma}$ converge.


\paragraph{Computational Complexity}
RANSAC-NN has a time complexity of $\mathcal{O}\big(snm \big)$ from ISP and $\mathcal{O}\big(tnm\big)$ from TS. 

\section{Experiments}
\label{sec:experiments}

We provide a comprehensive analysis of RANSAC-NN in the following. To begin, we conduct a benchmark of RANSAC-NN in comparison to other well-established OD methods in the setting where a clean inlier set is provided. Next, we consider the event where outlier contamination occurs in the training set, and we compare the performance difference caused by contaminated training. We then explore the use of RANSAC-NN for outlier removal in contaminated sets prior to OD model training. Lastly, we perform ablation studies to understand the components behind our algorithm.

\subsection{Experiment Setup}
\label{exp:experiment_setup}

\paragraph{Data and Metrics.}
For all our experiments, we follow the one-class classification setup \cite{2018_deepsvdd, one_class_svm, one_class_classification}. Benchmarks were conducted on natural image datasets SUN397 \cite{data-sun397}, Caltech101 \cite{data-caltech101}, and ImageNet21K\footnote{The original 1000 classes from ImageNet-1K were excluded.} \cite{data-imagenet21k}. 
For each dataset, we consider the top 10 classes with the most images, and we separate a train and test set for each class. We perform OD on each class, and samples from other classes are considered outliers. The ground truth labels from the datasets were adopted solely for performance evaluation. Following \cite{goad, 2021_exploring_limits_of_ood, 2019_mem_ae, knn-distance-algo, deecke2018anomaly}, we report the average ROC-AUC's from the 10 class evaluations. Each experiment is repeated 6 times to avoid over-fitting. 

\paragraph{Algorithms}
We evaluate RANSAC-NN alongside several well-established density estimation \cite{local_outlier_factor, algo-inne, isolation_forest, knn-distance-algo, 2018_deepsvdd, algo-lunar}, image reconstruction \cite{Outlier_Analysis_aggarwal, rca}, and self-supervised classification \cite{goad, 2021_neural_transformation} methods\footnote{Implemented by the PyOD \cite{zhao2019pyod} and DeepOD \cite{deepod} libraries.}. The RANSAC-NN sample size $m$ was set at $5\%$ of the dataset size for both ISP and TSP. The ISP sampling iterations $s$ was set at $\left \lceil{n/m}\right \rceil$ (each sample is expected to be sampled once). The TSP threshold iterations $t$ was kept constant at $500$.  

Prior to our experiments, we optimized the hyperparameters of every algorithm with $20$ automatic searches using HyperOpt \cite{hyperopt}. Image embeddings were extracted and normalized using a pre-trained ResNet-18 \cite{cnn-resnet} with the final fully-connected layer removed. Experiments on additional feature extractor choices and run-time performance have also been provided in the Appendix.

\paragraph{Terminology}
Outliers added to the training set are referred to as \textit{contamination}, and outliers added to the test set are referred to as \textit{perturbation}. Unless specified, it is assumed that the contamination level equals the perturbation level to ensure equal representation during evaluation. \textit{Algorithms} refer to the OD method, and \textit{models} are the results from training an algorithm on a dataset.

\subsection{Outlier Detection with Clean Training}
\label{exp:outlier_detection_benchmark}
In this experiment, we evaluate the performance of RANSAC-NN alongside several well-known OD algorithms. For each existing algorithm, we train its model on a \textbf{clean inlier set} and compare their performance under different levels of outlier perturbations. Since RANSAC-NN does not require training, we directly apply our algorithm to the perturbed test set.

\begin{table}[t]
    \centering
    \begin{adjustbox}{width=0.47\textwidth}
    \begin{tabular}{lllllc}
\toprule
\multirow{2}{*}{Algorithm} & \multicolumn{4}{c}{Outlier Perturbation (ROC)} & \multirow{2}{*}{Max $\sigma$}\\
\cmidrule(lr){2-5} 
{} &         $5\%$ &     $10\%$ &     $20\%$ &     $40\%$ & {} \\
\midrule
Isolation Forest \cite{isolation_forest} &      $0.966$ &  $0.964$ &  $0.963$ &  $0.960$ &  $\pm0.02$ \\
INNE \cite{algo-inne}             &      $0.973$ &  $0.976$ &  $0.976$ &  $0.970$ &  $\pm0.01$ \\
LOF  \cite{local_outlier_factor}            &      $0.977$ &  $0.976$ & $0.976$  &   $\boldsymbol{0.976}^{*}$ &  $\pm0.01$ \\
AutoEncoder  \cite{Outlier_Analysis_aggarwal}    &      $0.968$ &  $0.965$ &  $0.968$ &  $0.959$ &  $\pm0.02$ \\
LUNAR  \cite{algo-lunar}      &      $0.970$ &  $0.971$ &  $0.970$ &  $0.967$ &  $\pm0.02$ \\
KNN Distance \cite{knn-distance-algo}    &      $\boldsymbol{0.983}^{+}$ &  $\boldsymbol{0.984}^{+}$ &  $\boldsymbol{0.983}^{+}$ &  $\boldsymbol{0.981}^{+}$ &  $\pm0.01$ \\
GOAD \cite{goad}            &      $0.967$ &  $0.967$ &  $0.965$ &  $0.964$ &  $\pm0.02$ \\
Deep SVDD  \cite{2018_deepsvdd}      &      $0.922$ &  $0.926$ &  $0.919$ &  $0.913$ &  $\pm0.03$ \\
RCA  \cite{rca}            &      $0.922$ &  $0.918$ &  $0.915$ &  $0.895$ &  $\pm0.06$ \\
NeuTraL  \cite{2021_neural_transformation}  &      $0.931$ &  $0.936$ &  $0.931$ &  $0.926$ &  $\pm0.04$ \\
RANSAC-NN (Ours)      &      $\boldsymbol{0.980}^{*}$ &  $\boldsymbol{0.978}^{*}$ &  $\boldsymbol{0.977}^{*}$ &  $0.941$ &  $\pm0.04$ \\

\bottomrule
\end{tabular}
\end{adjustbox}

    \caption{\textbf{Outlier Detection with Clean Training.} Shown above are the average ROC-AUC's obtained by models trained using the corresponding algorithm on a clean set of inlier images. Notice that all models maintain a relatively solid performance despite the varying perturbation levels. This implies that existing algorithms are capable of image OD given they are trained with quality data. Although RANSAC-NN was directly applied to the test set without training, it still maintained favorable performance compared to the models that had undergone training.}
    \label{table:od_benchmark}
\end{table}

Table \ref{table:od_benchmark} shows the experiment results.
The first observation is that the majority of OD algorithms perform almost equally well when trained on an inlier set. For every algorithm, the performance differences are relatively minor under all perturbation levels. This implies that most existing algorithms are capable of maintaining a perturbation-robust image OD performance as long as a clean inlier set is available for training.

The second observation is that RANSAC-NN, despite the lack of training, can achieve similar performance to models that had undergone training. This is important as it entails the possibility of applying RANSAC-NN directly to a contaminated training set prior to the training of other OD models. In the next experiment, we demonstrate the importance of training on a clean inlier set by analyzing the impact of outlier contamination.

\subsection{Influence of Contaminated Training}
\label{exp:impact_of_contaminated_training}

\begin{figure}[t]
    \centering
    \includegraphics[width=0.47\textwidth]{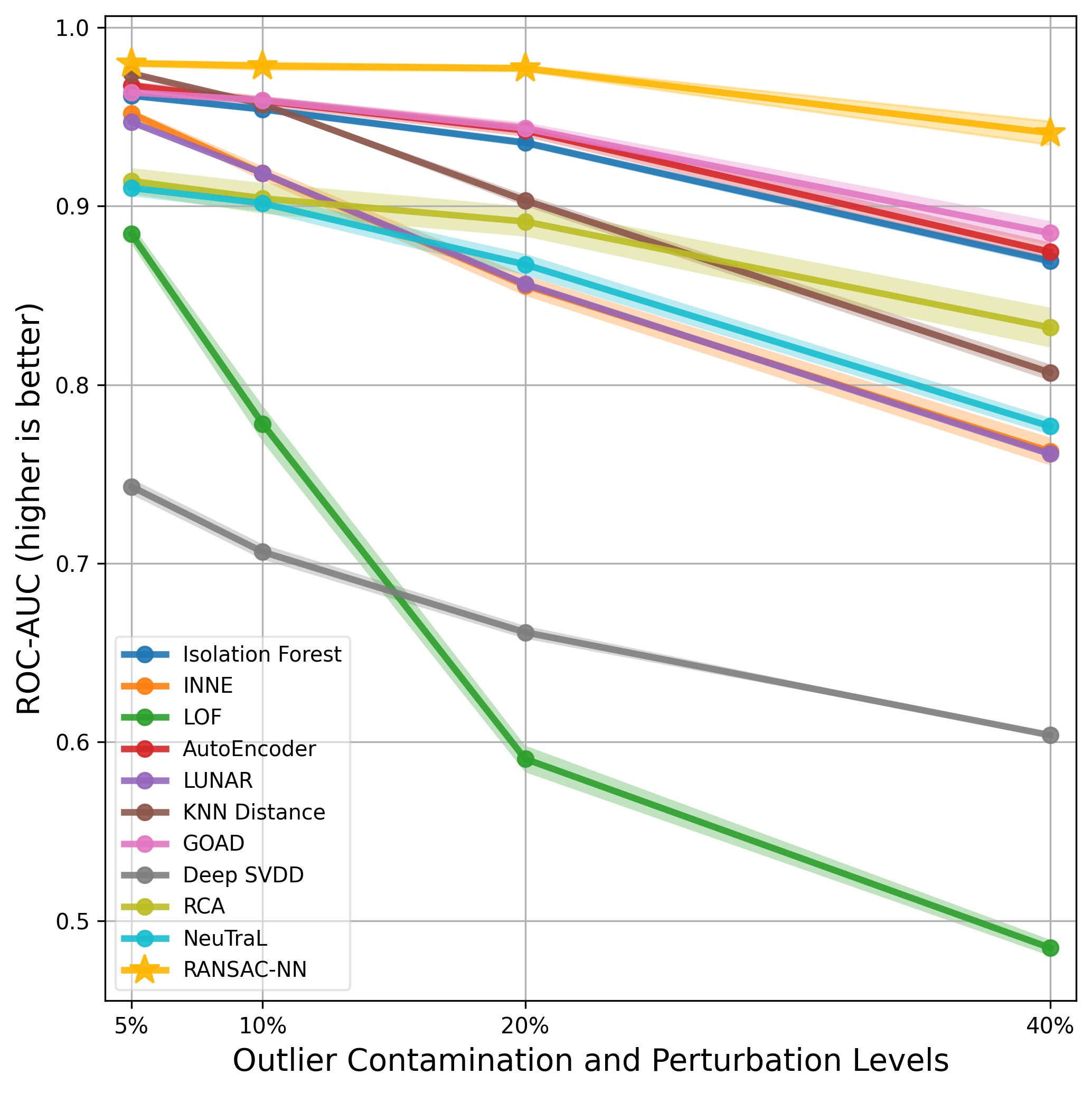}
    \caption{\textbf{Performance Drop from Contaminated Training.} Shown above are the performance of several OD algorithms when trained on a contaminated set. At low contamination levels, the drop in performance is already visible in some algorithms. As the contamination level increases, the performance drop is evident throughout all algorithms. In contrast to the uncontaminated setting, where the performance was relatively constant, this performance difference highlights the importance of maintaining a quality dataset when training OD models. Since RANSAC-NN does not require training, its performance is not influenced by contaminated training.
    }
    \label{fig:contaminated_training}
\end{figure}

In this experiment, we demonstrate the importance of maintaining an outlier-free dataset when training an OD model. We consider the same setup as in Section \ref{exp:outlier_detection_benchmark}, but we replace the clean training sets with \textbf{contaminated sets}. We then train a model for each algorithm at different contamination levels and compare their differences to the models that were trained on the clean sets. 

The performance difference due to outlier contamination is illustrated in Figure \ref{fig:contaminated_training}. We can observe that at a low contamination level, the performance drop is already noticeable in some algorithms. As the contamination level increases, the performance difference becomes more apparent. In comparison to the behavior of models that were trained on a clean inlier set (see Table \ref{table:od_benchmark}), this drop in performance further emphasizes the importance of maintaining a quality inlier set when training OD models.

Another observation is the relatively more robust performance delivered by RANSAC-NN under varying outlier perturbation levels. Since RANSAC-NN does not require training, its performance is not influenced by contamination in the training set. This observation suggests the possible use of RANSAC-NN as a filter to remove probable outliers from contaminated sets prior to model training.

\subsection{Outlier Filtering with RANSAC-NN}
\label{exp:outlier_filtering}

\begin{figure}[t]
    \centering
    \includegraphics[width=0.47\textwidth]{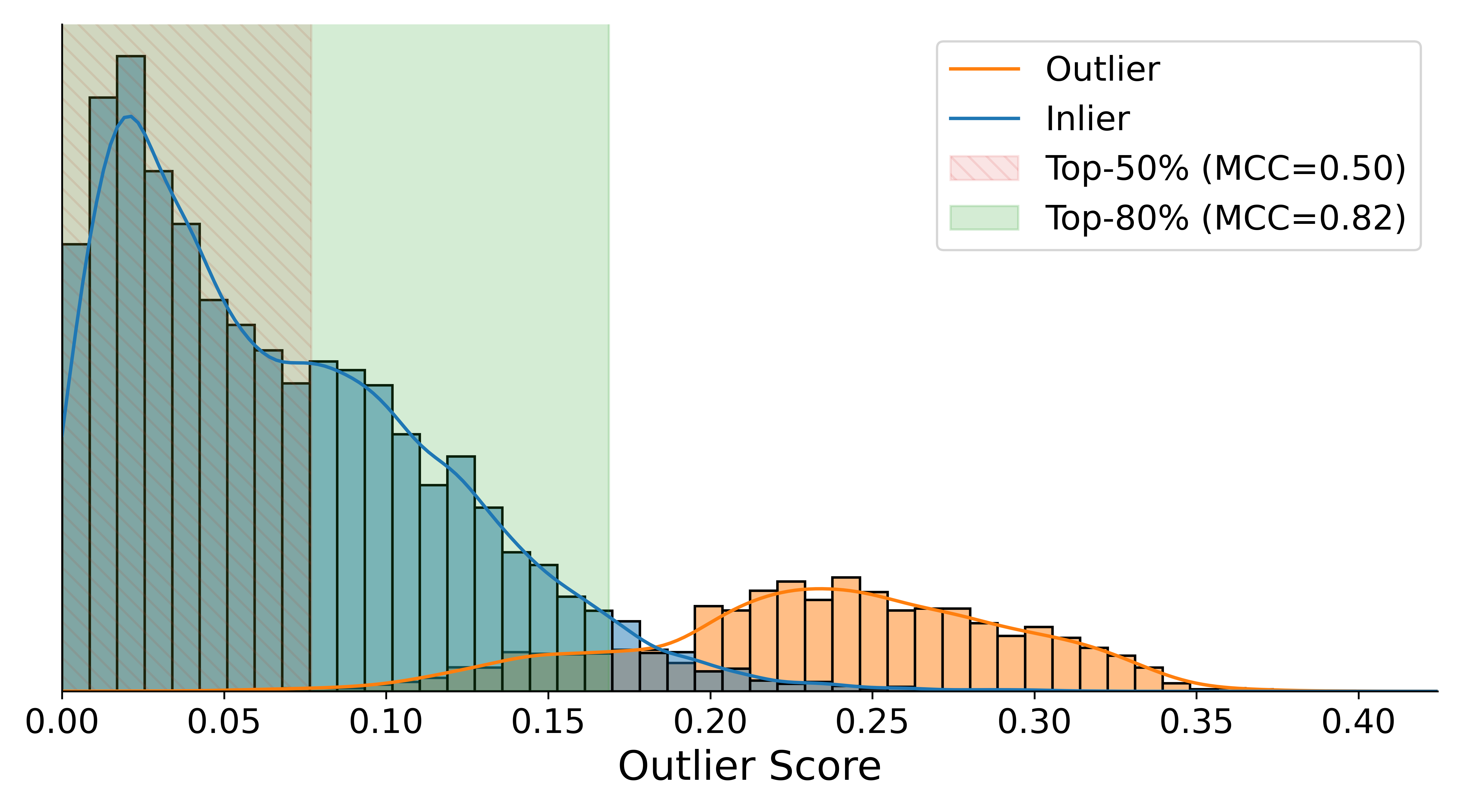}
    \caption{\textbf{Outlier Filtering with RANSAC-NN.} Shown above is the outlier score distribution predicted by RANSAC-NN on ImageNet21K \cite{data-imagenet21k} with $20\%$ perturbation. By taking the top-$p$ percent of images with the lowest outlier scores, we can easily remove large amounts of outliers. However, a smaller value of $p$ implies less inlier images available for training. In this example, $p=50$ yields lower selectivity than $p=80$ according to the Matthew's Correlation Coefficient \cite{mcc_advantage}.}
    \label{fig:outlier_filtering_dist}
\end{figure}

In this experiment, we assume that a contaminated set was provided, and we explore the use of outlier removal using RANSAC-NN. Before training an OD model for each algorithm, we applied RANSAC-NN on the contaminated set to obtain an outlier score distribution. Images were ranked according to their outlier scores in ascending order, and the top-$p$ percent of these images were selected for model training (see Figure~\ref{fig:outlier_filtering_dist}).
For each algorithm, a series of models were trained by taking different values of
$p$ (higher $p$ indicates more samples kept for training), and their results are detailed in Table~\ref{tab:outlier_filtering_results}.

\begin{table*}[t]
    \centering
    \begin{adjustbox}{width=\textwidth}
    \begin{tabular}{lllllll}
\toprule
\multirow{2}{*}{Algorithm} & \multicolumn{2}{c}{10\% Contamination} & \multicolumn{2}{c}{20\% Contamination} & \multicolumn{2}{c}{40\% Contamination}\\
\cmidrule(lr){2-3} \cmidrule(lr){4-5} \cmidrule(lr){6-7}
{} & Take Top-50\% & Take Top-90\% & Take Top-50\% & Take Top-80\% & Take Top-50\% & Take Top-60\%\\

\midrule
Isolation Forest \cite{isolation_forest} & $0.905\:(-4.23\%)$ & $\boldsymbol{0.959\:(+1.14\%)}$ & $0.921\:(-0.54\%)$ & $\boldsymbol{0.953\:(+2.69\%)}$ & $\boldsymbol{0.875\:(+7.98\%)}$ & $0.858\:(+6.29\%)$ \\ 
INNE \cite{algo-inne} & $0.855\:(-5.57\%)$ & $\boldsymbol{0.950\:(+3.98\%)}$ & $0.883\:(+3.58\%)$ & $\boldsymbol{0.930\:(+8.29\%)}$ & $0.862\:(+10.00\%)$ & $\boldsymbol{0.871\:(+10.91\%)}$ \\ 
LOF \cite{local_outlier_factor} & $0.869\:(+8.88\%)$ & $\boldsymbol{0.937\:(+15.69\%)}$ & $0.897\:(+32.14\%)$ & $\boldsymbol{0.904\:(+32.76\%)}$ & $\boldsymbol{0.869\:(+34.16\%)}$ & $0.853\:(+32.55\%)$ \\ 
AutoEncoder \cite{Outlier_Analysis_aggarwal} & $0.848\:(-11.02\%)$ & $\boldsymbol{0.961\:(+0.23\%)}$ & $0.874\:(-6.81\%)$ & $\boldsymbol{0.946\:(+0.35\%)}$ & $0.863\:(-1.11\%)$ & $\boldsymbol{0.891\:(+1.65\%)}$ \\ 
LUNAR \cite{algo-lunar} & $0.856\:(-4.70\%)$ & $\boldsymbol{0.953\:(+5.06\%)}$ & $0.887\:(+3.95\%)$ & $\boldsymbol{0.933\:(+8.51\%)}$ & $\boldsymbol{0.864\:(+14.01\%)}$ & $0.864\:(+14.00\%)$ \\ 
KNN Distance \cite{knn-distance-algo} & $\boldsymbol{0.977\:(+2.58\%)}$ & $0.976\:(+2.49\%)$ & $\boldsymbol{0.979\:(+7.80\%)}$ & $0.958\:(+5.76\%)$ & $\boldsymbol{0.900\:(+14.07\%)}$ & $0.891\:(+13.12\%)$ \\ 
GOAD \cite{goad} & $\boldsymbol{0.966\:(+0.69\%)}$ & $0.964\:(+0.47\%)$ & $\boldsymbol{0.963\:(+1.94\%)}$ & $0.956\:(+1.31\%)$ & $\boldsymbol{0.939\:(+5.39\%)}$ & $0.930\:(+4.53\%)$ \\ 
Deep SVDD \cite{2018_deepsvdd} & $\boldsymbol{0.950\:(+24.67\%)}$ & $0.873\:(+17.05\%)$ & $\boldsymbol{0.942\:(+28.24\%)}$ & $0.860\:(+19.96\%)$ & $\boldsymbol{0.861\:(+25.54\%)}$ & $0.825\:(+22.00\%)$ \\ 
RCA \cite{rca} & $0.896\:(-0.82\%)$ & $\boldsymbol{0.908\:(+0.35\%)}$ & $\boldsymbol{0.901\:(+0.94\%)}$ & $0.901\:(+0.94\%)$ & $\boldsymbol{0.868\:(+3.59\%)}$ & $0.862\:(+2.99\%)$ \\
NeuTraL \cite{2021_neural_transformation} & $\boldsymbol{0.933\:(+3.19\%)}$ & $0.922\:(+2.09\%)$ & $\boldsymbol{0.925\:(+5.77\%)}$ & $0.905\:(+3.77\%)$ & $\boldsymbol{0.881\:(+10.48\%)}$ & $0.867\:(+9.00\%)$ \\ 
\bottomrule
\end{tabular}
\end{adjustbox}
    \caption{\textbf{Outlier Filtering with RANSAC-NN.} Shown above are the performance of OD models trained on a RANSAC-NN filtered dataset (alongside their relative difference to contaminated training). The datasets were initially contaminated, and we applied RANSAC-NN to generate outlier scores for each training image. We then took the top-$p$ percent of images with the lowest outlier scores and trained the models using the filtered set. As demonstrated above, filtering the dataset using RANSAC-NN has led to significant performance gains in most OD algorithms, especially in the high contamination setting.}
    \label{tab:outlier_filtering_results}
\end{table*}

From the results, it is evident that models trained with a filtered set have resulted in significant performance gains compared to those trained directly on the contaminated set. At $10\%$ contamination, we can notice that the majority of the algorithms benefit more by removing fewer samples and keeping more for training. At a higher contamination of $40\%$, almost all algorithms experience a significantly better performance by removing more samples. Essentially, in cases where contamination in the training set is known to be severe, higher precision should favored for better OD performance during the outlier removal process.

Another interesting observation from this experiment is the characteristics of the OD algorithms. INNE \cite{algo-inne} and AutoEncoder \cite{Outlier_Analysis_aggarwal}, in particular, are much more dependent on the amount of training data. In each contaminated setting, both algorithms experienced better improvements by selecting a tighter threshold closer to the actual contamination level. In contrast, Deep SVDD \cite{2018_deepsvdd}, GOAD \cite{goad}, NeuTraL\cite{2021_neural_transformation}, and KNN Distance \cite{knn-distance-algo} were able to deliver better performance with less but potentially higher quality data. 

\subsection{Ablation Studies}
\label{exp:ablation_studies}

\begin{figure}[t]
    \centering
    \includegraphics[width=0.47\textwidth]{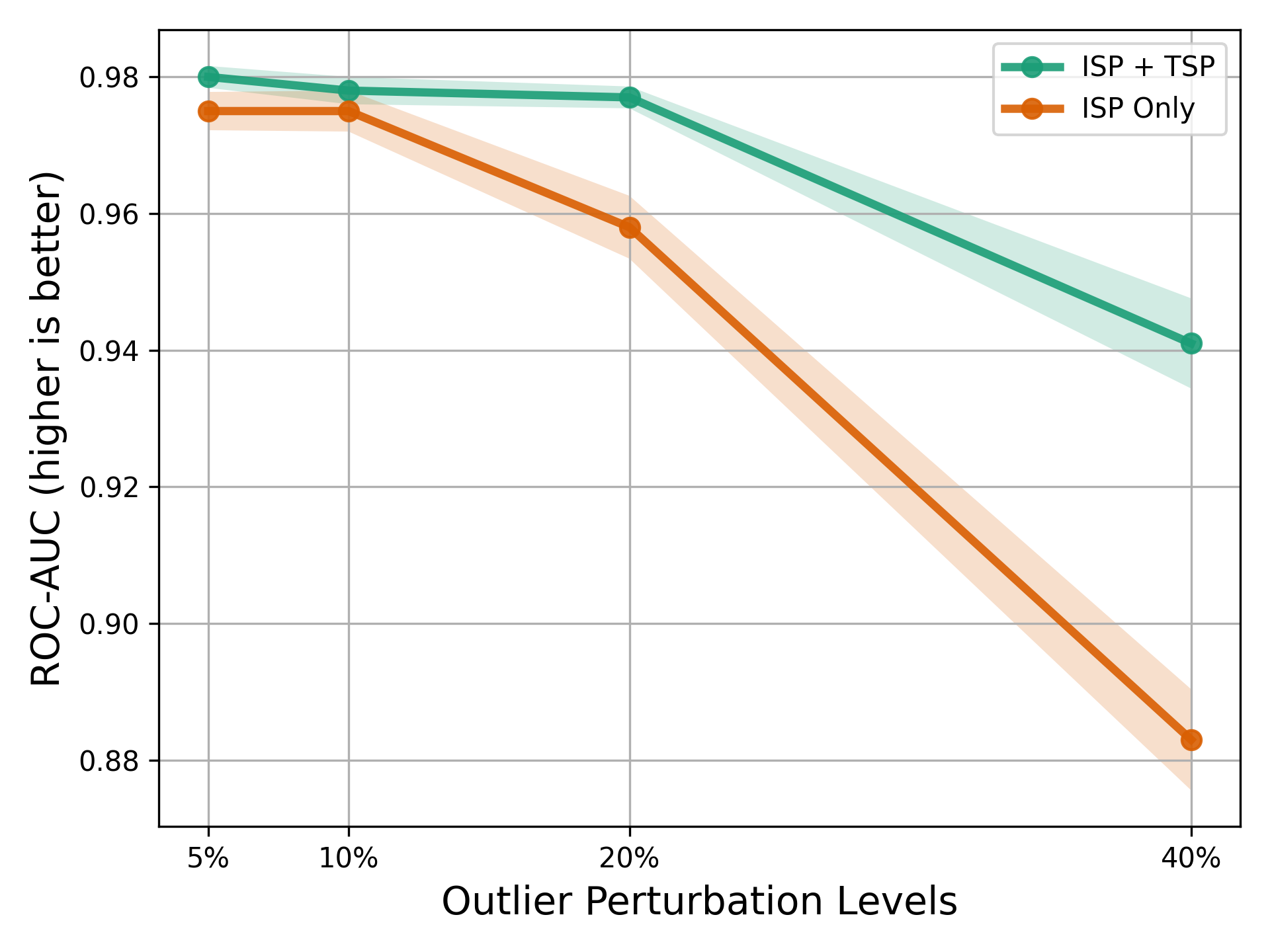}
    \caption{\textbf{Improvements from Threshold Sampling.} Plotted above is the performance improvements from applying Threshold Sampling (TS) in comparison to using the inverted inlier scores from ISP. Notice how applying TS improves the robustness of RANSAC-NN especially in high perturbation settings. 
    }
    \label{fig:threshold-sampling-improvement}
\end{figure}

\paragraph{Improvements from Threshold Sampling}
In this experiment, we analyze the performance improvements contributed by Threshold Sampling (TS). We invert the inlier scores predicted by ISP and compare them to the outlier scores produced by TS. In Figure \ref{fig:threshold-sampling-improvement}, we illustrate the performance obtained under both setups with varying outlier perturbation levels. From the results, we can observe that TS indeed enhances the predictive performance of RANSAC-NN. This is especially true when outlier perturbation is severe, indicating the importance of TS in increasing the robustness of RANSAC-NN. 

\paragraph{Sample Size and Sampling Iteration}
\label{exp:hyperparameters}

The two primary hyperparameters of RANSAC-NN are the sample size $m$ and the sampling iterations $s$. As described in Section \ref{methods:sample_size_iteration_properties}, the sample size $m$ influences the probability of sampling a clean inlier set, and the sampling iteration $s$ impacts the odds of sampling at least one clean inlier set. In this experiment, we compare the performance of RANSAC-NN under different sample sizes and sampling iteration configurations. 

\begin{figure}[t]
    \centering
    \includegraphics[width=0.47\textwidth]{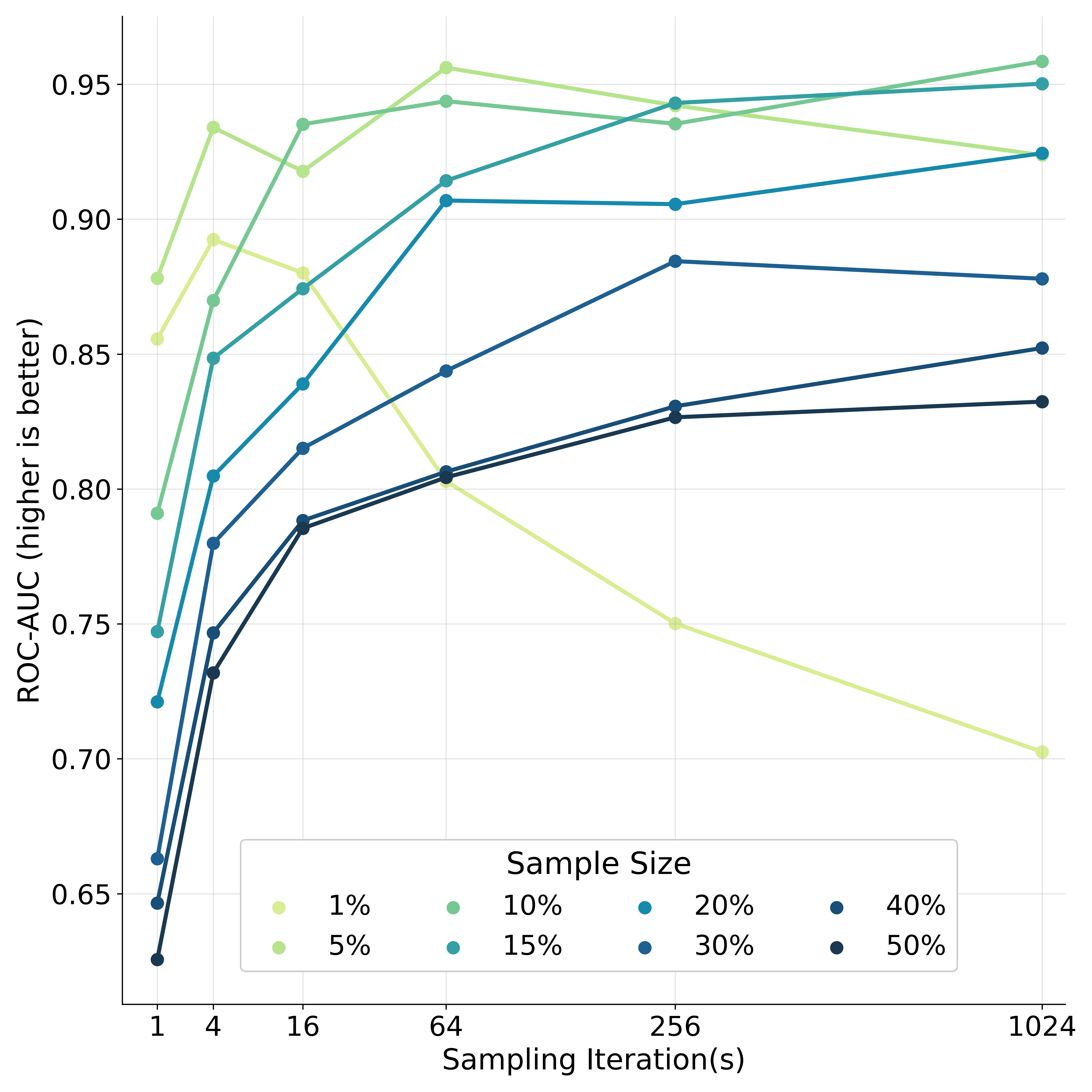}
    \caption{
    \textbf{Sample Size and Sampling Iterations.} Shown above are the RANSAC-NN performance under different sample sizes (color intensity) and sampling iterations ($x$-axis). Notice how a large sample size requires significantly more sampling iterations to achieve similar performance obtained by smaller sample sizes. However, an extremely small sample size should be avoided (see Section \ref{methods:sample_size_iteration_properties}). With a reasonable sample size, larger sampling iterations often result in better performance.
    }
    \label{fig:sample_size_iteration}
\end{figure}

From the results in Figure \ref{fig:sample_size_iteration}, we can observe that an extreme sample size causes unstable performance, especially as the sampling iteration increases. This is because a small enough sample size increases the odds of sampling a set containing all outliers. With the addition of repeated sampling, this probability only increases, which is why we observe such unfavorable performance with sample size at $1\%$ of the dataset. 

On the other extreme, a large sample size reduces the probability of sampling an all-inlier set. This phenomenon can be seen in cases where the sample size exceeds $30\%$ of the dataset. The ideal sample size, therefore, falls within the $10-20\%$ range. With the addition of large sampling iterations, the performance continues improving as we increase the odds of sampling an all-inlier set. 

\section{Conclusions and Future Work}
\label{sec:conclusions}

This paper explores the problem of outlier contamination during the development of image OD models. We discuss the strengths and weaknesses of existing algorithms by comparing their performance when trained on different datasets. In our experiments, we observe solid performance in all existing methods when trained on clean inlier images. The performance of these algorithms, however, declines when outliers images contaminate the training data. This issue limits the use of existing OD methods on new image domains as datasets must be carefully examined before undergoing training.

This paper also introduces an alternate image OD algorithm called RANSAC-NN that avoids the need of data preparation or model training. Unlike existing methods, whose objective is to train a task-specific OD model for inference, the goal of RANSAC-NN is to detect outlier in contaminated datasets using the data distribution alone. 
Despite the lack of training, RANSAC-NN remains competitive against existing methods on a range of benchmarks. Furthermore, we demonstrate the use of RANSAC-NN to increase the robustness of existing OD methods when training on data containing potential outliers. Additional ablation studies on the components of RANSAC-NN have also been explored. 





Future works may include extending RANSAC-NN into other domains such as text or audio. Possible applications in image mislabeled detection may also be explored (see Appendix). With the findings thus far, we hope RANSAC-NN may be a valuable tool for future image OD applications.

{
    \small
    \bibliographystyle{ieeenat_fullname}
    \bibliography{main}

\begin{thebibliography}{55}
\providecommand{\natexlab}[1]{#1}
\providecommand{\url}[1]{\texttt{#1}}
\expandafter\ifx\csname urlstyle\endcsname\relax
  \providecommand{\doi}[1]{doi: #1}\else
  \providecommand{\doi}{doi: \begingroup \urlstyle{rm}\Url}\fi

\bibitem[Abati et~al.(2019)Abati, Porrello, Calderara, and Cucchiara]{2019_as_ae}
Davide Abati, Angelo Porrello, Simone Calderara, and Rita Cucchiara.
\newblock Latent space autoregression for novelty detection.
\newblock In \emph{Proceedings of the IEEE/CVF conference on computer vision and pattern recognition}, pages 481--490, 2019.

\bibitem[Aggarwal(2016)]{Outlier_Analysis_aggarwal}
Charu~C. Aggarwal.
\newblock \emph{Outlier Analysis}.
\newblock Springer Publishing Company, Incorporated, 2nd edition, 2016.

\bibitem[Ali et~al.(2020)Ali, Khan, and Kyung]{2020_ssl_ae}
Rabia Ali, Muhammad Umar~Karim Khan, and Chong~Min Kyung.
\newblock Self-supervised representation learning for visual anomaly detection.
\newblock \emph{arXiv preprint arXiv:2006.09654}, 2020.

\bibitem[An and Cho(2015)]{2015_vae_anomaly_det}
Jinwon An and Sungzoon Cho.
\newblock Variational autoencoder based anomaly detection using reconstruction probability.
\newblock \emph{Special lecture on IE}, 2\penalty0 (1):\penalty0 1--18, 2015.

\bibitem[Balestriero et~al.(2023)Balestriero, Ibrahim, Sobal, Morcos, Shekhar, Goldstein, Bordes, Bardes, Mialon, Tian, Schwarzschild, Wilson, Geiping, Garrido, Fernandez, Bar, Pirsiavash, LeCun, and Goldblum]{ssl_cookbook}
Randall Balestriero, Mark Ibrahim, Vlad Sobal, Ari~S. Morcos, Shashank Shekhar, Tom Goldstein, Florian Bordes, Adrien Bardes, Gr{\'e}goire Mialon, Yuandong Tian, Avi Schwarzschild, Andrew~Gordon Wilson, Jonas Geiping, Quentin Garrido, Pierre Fernandez, Amir Bar, Hamed Pirsiavash, Yann LeCun, and Micah Goldblum.
\newblock A cookbook of self-supervised learning.
\newblock \emph{ArXiv}, abs/2304.12210, 2023.

\bibitem[Bandaragoda et~al.(2014)Bandaragoda, Ting, Albrecht, Liu, and Wells]{algo-inne}
Tharindu~R. Bandaragoda, Kai~Ming Ting, David Albrecht, Fei~Tony Liu, and Jonathan~R. Wells.
\newblock Efficient anomaly detection by isolation using nearest neighbour ensemble.
\newblock In \emph{2014 IEEE International Conference on Data Mining Workshop}, pages 698--705, 2014.

\bibitem[Bergman and Hoshen(2020)]{goad}
Liron Bergman and Yedid Hoshen.
\newblock Classification-based anomaly detection for general data.
\newblock In \emph{International Conference on Learning Representations (ICLR)}, 2020.

\bibitem[Bergstra et~al.(2013)Bergstra, Yamins, and Cox]{hyperopt}
James Bergstra, Daniel Yamins, and David~D. Cox.
\newblock Making a science of model search: Hyperparameter optimization in hundreds of dimensions for vision architectures.
\newblock In \emph{International Conference on Machine Learning}, 2013.

\bibitem[Bishop(2006)]{book-prml-bishop}
Christopher~M. Bishop.
\newblock \emph{Pattern Recognition and Machine Learning (Information Science and Statistics)}.
\newblock Springer-Verlag, Berlin, Heidelberg, 2006.

\bibitem[Breunig et~al.(2000)Breunig, Kriegel, Ng, and Sander]{local_outlier_factor}
Markus~M. Breunig, Hans-Peter Kriegel, Raymond~T. Ng, and J\"{o}rg Sander.
\newblock Lof: Identifying density-based local outliers.
\newblock \emph{SIGMOD Rec.}, 29\penalty0 (2):\penalty0 93–104, 2000.

\bibitem[Chicco and Jurman(2020)]{mcc_advantage}
Davide Chicco and Giuseppe Jurman.
\newblock The advantages of the matthews correlation coefficient (mcc) over f1 score and accuracy in binary classification evaluation.
\newblock \emph{BMC Genomics}, 21, 2020.

\bibitem[Deecke et~al.(2018)Deecke, Vandermeulen, Ruff, Mandt, and Kloft]{deecke2018anomaly}
Lucas Deecke, Robert Vandermeulen, Lukas Ruff, Stephan Mandt, and Marius Kloft.
\newblock Anomaly detection with generative adversarial networks, 2018.

\bibitem[Erfani et~al.(2016)Erfani, Rajasegarar, Karunasekera, and Leckie]{one_class_svm}
Sarah~M. Erfani, Sutharshan Rajasegarar, Shanika Karunasekera, and Christopher Leckie.
\newblock High-dimensional and large-scale anomaly detection using a linear one-class svm with deep learning.
\newblock \emph{Pattern Recognition}, 58:\penalty0 121--134, 2016.

\bibitem[Fei-Fei et~al.(2004)Fei-Fei, Fergus, and Perona]{data-caltech101}
Li Fei-Fei, Rob Fergus, and Pietro Perona.
\newblock Learning generative visual models from few training examples: An incremental bayesian approach tested on 101 object categories.
\newblock \emph{Computer Vision and Pattern Recognition Workshop}, 2004.

\bibitem[Fischler and Bolles(1981)]{ref_ransac}
Martin~A. Fischler and Robert~C. Bolles.
\newblock Random sample consensus: A paradigm for model fitting with applications to image analysis and automated cartography.
\newblock \emph{Commun. ACM}, 24\penalty0 (6):\penalty0 381–395, 1981.

\bibitem[Fort et~al.(2021)Fort, Ren, and Lakshminarayanan]{2021_exploring_limits_of_ood}
Stanislav Fort, Jie Ren, and Balaji Lakshminarayanan.
\newblock Exploring the limits of out-of-distribution detection.
\newblock In \emph{Advances in Neural Information Processing Systems}, pages 7068--7081. Curran Associates, Inc., 2021.

\bibitem[Golan and El-Yaniv(2018)]{geom}
Izhak Golan and Ran El-Yaniv.
\newblock Deep anomaly detection using geometric transformations.
\newblock In \emph{Advances in Neural Information Processing Systems}. Curran Associates, Inc., 2018.

\bibitem[Gong et~al.(2019)Gong, Liu, Le, Saha, Mansour, Venkatesh, and Hengel]{2019_mem_ae}
Dong Gong, Lingqiao Liu, Vuong Le, Budhaditya Saha, Moussa~Reda Mansour, Svetha Venkatesh, and Anton van~den Hengel.
\newblock Memorizing normality to detect anomaly: Memory-augmented deep autoencoder for unsupervised anomaly detection.
\newblock In \emph{Proceedings of the IEEE/CVF International Conference on Computer Vision}, pages 1705--1714, 2019.

\bibitem[Goodfellow et~al.(2014)Goodfellow, Pouget-Abadie, Mirza, Xu, Warde-Farley, Ozair, Courville, and Bengio]{2014_gans}
Ian~J. Goodfellow, Jean Pouget-Abadie, Mehdi Mirza, Bing Xu, David Warde-Farley, Sherjil Ozair, Aaron Courville, and Yoshua Bengio.
\newblock Generative adversarial nets.
\newblock In \emph{Proceedings of the 27th International Conference on Neural Information Processing Systems - Volume 2}, page 2672–2680, Cambridge, MA, USA, 2014. MIT Press.

\bibitem[Goodge et~al.(2022)Goodge, Hooi, Ng, and Ng]{algo-lunar}
Adam Goodge, Bryan Hooi, See~Kiong Ng, and Wee~Siong Ng.
\newblock Lunar: Unifying local outlier detection methods via graph neural networks.
\newblock 2022.

\bibitem[He et~al.(2016)He, Zhang, Ren, and Sun]{cnn-resnet}
Kaiming He, Xiangyu Zhang, Shaoqing Ren, and Jian Sun.
\newblock Deep residual learning for image recognition.
\newblock In \emph{2016 IEEE Conference on Computer Vision and Pattern Recognition (CVPR)}, pages 770--778, 2016.

\bibitem[Hendrycks et~al.(2019)Hendrycks, Mazeika, Kadavath, and Song]{2019_using_ssl_can_improve_robustness}
Dan Hendrycks, Mantas Mazeika, Saurav Kadavath, and Dawn Song.
\newblock Using self-supervised learning can improve model robustness and uncertainty.
\newblock \emph{Advances in Neural Information Processing Systems (NeurIPS)}, 2019.

\bibitem[Hodge and Austin(2004)]{od_survey}
Victoria Hodge and Jim Austin.
\newblock A survey of outlier detection methodologies.
\newblock \emph{Artificial intelligence review}, 22:\penalty0 85--126, 2004.

\bibitem[Howard et~al.(2019)Howard, Sandler, Chu, Chen, Chen, Tan, Wang, Zhu, Pang, Vasudevan, Le, and Adam]{mobilenet-v3}
Andrew~G. Howard, Mark Sandler, Grace Chu, Liang-Chieh Chen, Bo Chen, Mingxing Tan, Weijun Wang, Yukun Zhu, Ruoming Pang, Vijay Vasudevan, Quoc~V. Le, and Hartwig Adam.
\newblock Searching for mobilenetv3.
\newblock \emph{2019 IEEE/CVF International Conference on Computer Vision (ICCV)}, pages 1314--1324, 2019.

\bibitem[Johnson et~al.(2019)Johnson, Douze, and J{\'e}gou]{faiss}
Jeff Johnson, Matthijs Douze, and Herv{\'e} J{\'e}gou.
\newblock Billion-scale similarity search with {GPUs}.
\newblock \emph{IEEE Transactions on Big Data}, 7\penalty0 (3):\penalty0 535--547, 2019.

\bibitem[Khan and Madden(2010)]{a_surve_of_one_class_khan}
Shehroz~S. Khan and Michael~G. Madden.
\newblock A survey of recent trends in one class classification.
\newblock In \emph{Artificial Intelligence and Cognitive Science}, pages 188--197, Berlin, Heidelberg, 2010. Springer Berlin Heidelberg.

\bibitem[Kuan and Mueller(2022)]{knn-distance-algo}
Johnson Kuan and Jonas Mueller.
\newblock Back-to-the-basics: Revisiting out-of-distribution detection baselines.
\newblock In \emph{Proceedings of the International Conference on Machine Learning (ICML) Workshops, Workshop on Principles of Distribution Shift}, 2022.

\bibitem[Liu et~al.(2021)Liu, Wang, Lin, Tan, and Zhou]{rca}
Boyang Liu, Ding Wang, Kaixiang Lin, Pang-Ning Tan, and Jiayu Zhou.
\newblock Rca: A deep collaborative autoencoder approach for anomaly detection.
\newblock In \emph{Proceedings of the Thirtieth International Joint Conference on Artificial Intelligence, {IJCAI-21}}, pages 1505--1511. International Joint Conferences on Artificial Intelligence Organization, 2021.
\newblock Main Track.

\bibitem[Liu et~al.(2008)Liu, Ting, and Zhou]{isolation_forest}
Fei~Tony Liu, Kai~Ming Ting, and Zhi-Hua Zhou.
\newblock Isolation forest.
\newblock In \emph{2008 Eighth IEEE International Conference on Data Mining}, pages 413--422, 2008.

\bibitem[Liu et~al.(2018)Liu, Li, Zhou, Jiang, Sun, Wang, and He]{sogann}
Yezheng Liu, Zhe Li, Chong Zhou, Yuanchun Jiang, Jianshan Sun, M. Wang, and Xiangnan He.
\newblock Generative adversarial active learning for unsupervised outlier detection.
\newblock \emph{IEEE Transactions on Knowledge and Data Engineering}, 32:\penalty0 1517--1528, 2018.

\bibitem[Luan et~al.(2021)Luan, Gu, Freidovich, Jiang, and Zhao]{iForest_on_NN}
Siyu Luan, Zonghua Gu, Leonid~B. Freidovich, Lili Jiang, and Qingling Zhao.
\newblock Out-of-distribution detection for deep neural networks with isolation forest and local outlier factor.
\newblock \emph{IEEE Access}, 9:\penalty0 132980--132989, 2021.

\bibitem[Moya et~al.(1993)Moya, Koch, and Hostetler]{one_class_classification}
M~M Moya, M~W Koch, and L~D Hostetler.
\newblock One-class classifier networks for target recognition applications.
\newblock In \emph{Proceedings World Congress on Neural Networks}, 1993.

\bibitem[Oza and Patel(2018)]{2018_oneclass_cnn}
Poojan Oza and Vishal~M Patel.
\newblock One-class convolutional neural network.
\newblock \emph{IEEE Signal Processing Letters}, 26\penalty0 (2):\penalty0 277--281, 2018.

\bibitem[Perera and Patel(2019)]{2019_deep_feature_for_oneclass_cls}
Pramuditha Perera and Vishal~M Patel.
\newblock Learning deep features for one-class classification.
\newblock \emph{IEEE Transactions on Image Processing}, 28\penalty0 (11):\penalty0 5450--5463, 2019.

\bibitem[Pevn{\'y}(2016)]{loda}
Tom{\'a}{\vs} Pevn{\'y}.
\newblock Loda: Lightweight on-line detector of anomalies.
\newblock \emph{Machine Learning}, 102:\penalty0 275--304, 2016.

\bibitem[Qiu et~al.(2021)Qiu, Pfrommer, Kloft, Mandt, and Rudolph]{2021_neural_transformation}
Chen Qiu, Timo Pfrommer, Marius Kloft, Stephan Mandt, and Maja Rudolph.
\newblock Neural transformation learning for deep anomaly detection beyond images.
\newblock In \emph{International Conference on Machine Learning}, pages 8703--8714. PMLR, 2021.

\bibitem[Ren et~al.(2015)Ren, He, Girshick, and Sun]{fasterrcnn}
Shaoqing Ren, Kaiming He, Ross Girshick, and Jian Sun.
\newblock Faster {R-CNN}: Towards real-time object detection with region proposal networks.
\newblock In \emph{Advances in Neural Information Processing Systems ({NIPS})}, 2015.

\bibitem[Ridnik et~al.(2021)Ridnik, Ben-Baruch, Noy, and Zelnik]{data-imagenet21k}
Tal Ridnik, Emanuel Ben-Baruch, Asaf Noy, and Lihi Zelnik.
\newblock Imagenet-21k pretraining for the masses.
\newblock In \emph{Proceedings of the Neural Information Processing Systems Track on Datasets and Benchmarks}. Curran, 2021.

\bibitem[Ruff et~al.(2018)Ruff, Vandermeulen, Goernitz, Deecke, Siddiqui, Binder, M{\"u}ller, and Kloft]{2018_deepsvdd}
Lukas Ruff, Robert Vandermeulen, Nico Goernitz, Lucas Deecke, Shoaib~Ahmed Siddiqui, Alexander Binder, Emmanuel M{\"u}ller, and Marius Kloft.
\newblock Deep one-class classification.
\newblock In \emph{Proceedings of the 35th International Conference on Machine Learning}, pages 4393--4402. PMLR, 2018.

\bibitem[Sakurada and Yairi(2014)]{2014_anomaly_detection}
Mayu Sakurada and Takehisa Yairi.
\newblock Anomaly detection using autoencoders with nonlinear dimensionality reduction.
\newblock In \emph{Proceedings of the MLSDA 2014 2nd workshop on machine learning for sensory data analysis}, pages 4--11, 2014.

\bibitem[Salehi et~al.(2020)Salehi, Eftekhar, Sadjadi, Rohban, and Rabiee]{2020_puzzle_ae}
Mohammadreza Salehi, Ainaz Eftekhar, Niousha Sadjadi, Mohammad~Hossein Rohban, and Hamid~R Rabiee.
\newblock Puzzle-ae: Novelty detection in images through solving puzzles.
\newblock \emph{arXiv preprint arXiv:2008.12959}, 2020.

\bibitem[Schlegl et~al.(2017)Schlegl, Seeb{\"o}ck, Waldstein, Schmidt-Erfurth, and Langs]{anogan}
Thomas Schlegl, Philipp Seeb{\"o}ck, Sebastian~M. Waldstein, Ursula Schmidt-Erfurth, and Georg Langs.
\newblock Unsupervised anomaly detection with generative adversarial networks to guide marker discovery.
\newblock In \emph{Information Processing in Medical Imaging}, pages 146--157, Cham, 2017. Springer International Publishing.

\bibitem[Schölkopf et~al.(2001)Schölkopf, Platt, Shawe-Taylor, Smola, and Williamson]{oc-svm}
Bernhard Schölkopf, John~C. Platt, John Shawe-Taylor, Alex~J. Smola, and Robert~C. Williamson.
\newblock {Estimating the Support of a High-Dimensional Distribution}.
\newblock \emph{Neural Computation}, 13\penalty0 (7):\penalty0 1443--1471, 2001.

\bibitem[Sehwag et~al.(2021)Sehwag, Chiang, and Mittal]{2021_ssd_od}
Vikash Sehwag, Mung Chiang, and Prateek Mittal.
\newblock Ssd: A unified framework for self-supervised outlier detection.
\newblock \emph{arXiv preprint arXiv:2103.12051}, 2021.

\bibitem[Tack et~al.(2020)Tack, Mo, Jeong, and Shin]{2020_csi_ssl_ae}
Jihoon Tack, Sangwoo Mo, Jongheon Jeong, and Jinwoo Shin.
\newblock Csi: Novelty detection via contrastive learning on distributionally shifted instances.
\newblock \emph{Advances in neural information processing systems}, 33:\penalty0 11839--11852, 2020.

\bibitem[Tax and Duin(2004)]{2004_svdd}
David~MJ Tax and Robert~PW Duin.
\newblock Support vector data description.
\newblock \emph{Machine learning}, 54:\penalty0 45--66, 2004.

\bibitem[Vasu et~al.(2023)Vasu, Gabriel, Zhu, Tuzel, and Ranjan]{fastvit}
Pavan Kumar~Anasosalu Vasu, James Gabriel, Jeff Zhu, Oncel Tuzel, and Anurag Ranjan.
\newblock Fastvit: A fast hybrid vision transformer using structural reparameterization.
\newblock In \emph{Proceedings of the IEEE/CVF International Conference on Computer Vision}, 2023.

\bibitem[Xia et~al.(2015)Xia, Cao, Wen, Hua, and Sun]{2015_autoencoder_reconstruction_loss}
Yan Xia, Xudong Cao, Fang Wen, Gang Hua, and Jian Sun.
\newblock Learning discriminative reconstructions for unsupervised outlier removal.
\newblock In \emph{2015 IEEE International Conference on Computer Vision (ICCV)}, pages 1511--1519, 2015.

\bibitem[{Xiao} et~al.(2010){Xiao}, {Hays}, {Ehinger}, {Oliva}, and {Torralba}]{data-sun397}
J. {Xiao}, J. {Hays}, K.~A. {Ehinger}, A. {Oliva}, and A. {Torralba}.
\newblock Sun database: Large-scale scene recognition from abbey to zoo.
\newblock In \emph{2010 IEEE Computer Society Conference on Computer Vision and Pattern Recognition}, pages 3485--3492, 2010.

\bibitem[Xu et~al.(2023)Xu, Pang, Wang, and Wang]{deepod}
Hongzuo Xu, Guansong Pang, Yijie Wang, and Yongjun Wang.
\newblock Deep isolation forest for anomaly detection.
\newblock \emph{IEEE Transactions on Knowledge and Data Engineering}, pages 1--14, 2023.

\bibitem[Yang et~al.(2021{\natexlab{a}})Yang, Xu, Qi, and Shi]{ImageOD-survey}
Jie Yang, Ruijie Xu, Zhiquan Qi, and Yong Shi.
\newblock Visual anomaly detection for images: A survey.
\newblock \emph{arXiv preprint arXiv:2109.13157}, 2021{\natexlab{a}}.

\bibitem[Yang et~al.(2021{\natexlab{b}})Yang, Zhou, Li, and Liu]{generalize_ood_survey}
Jingkang Yang, Kaiyang Zhou, Yixuan Li, and Ziwei Liu.
\newblock Generalized out-of-distribution detection: A survey.
\newblock \emph{arXiv preprint arXiv:2110.11334}, 2021{\natexlab{b}}.

\bibitem[Ye et~al.(2020)Ye, Huang, Cao, Li, Zhang, and Lu]{2020_attr}
Fei Ye, Chaoqin Huang, Jinkun Cao, Maosen Li, Ya Zhang, and Cewu Lu.
\newblock Attribute restoration framework for anomaly detection.
\newblock \emph{IEEE Transactions on Multimedia}, 24:\penalty0 116--127, 2020.

\bibitem[Zhao et~al.(2019)Zhao, Nasrullah, and Li]{zhao2019pyod}
Yue Zhao, Zain Nasrullah, and Zheng Li.
\newblock Pyod: A python toolbox for scalable outlier detection.
\newblock \emph{Journal of Machine Learning Research}, 20\penalty0 (96):\penalty0 1--7, 2019.

\bibitem[Zong et~al.(2018)Zong, Song, Min, Cheng, Lumezanu, Cho, and Chen]{2018_dae_gmm}
Bo Zong, Qi Song, Martin~Renqiang Min, Wei Cheng, Cristian Lumezanu, Daeki Cho, and Haifeng Chen.
\newblock Deep autoencoding gaussian mixture model for unsupervised anomaly detection.
\newblock In \emph{International conference on learning representations}, 2018.

\end{thebibliography}
}

\clearpage
\setcounter{page}{1}
\maketitlesupplementary

\section{Variations in Feature Extractors}

We include additional experiments using MobileNet-v3 \cite{mobilenet-v3} and Fast-ViT \cite{fastvit} as the feature extractor of choice. We follow the same experiment setup in Section \ref{sec:experiments}, and the results are presented in the following. 

\subsection{Outlier Detection with Clean Training}
\label{suppl:clean_training_benchmark}

The same experiments from Section \ref{exp:outlier_detection_benchmark} are repeated using MobileNet-v3 and Fast-ViT, and their results are listed in Tables \ref{suppl:clean_train_mobilenet} and \ref{suppl:clean_train_swin}. As demonstrated below, the performance of models trained on a \textbf{clean} inlier set remain consistent regardless of the choice in feature extractor.

\begin{table}[h]
    \centering
    \begin{adjustbox}{width=0.47\textwidth}
    \begin{tabular}{lllllc}
\toprule
\multirow{2}{*}{Algorithm} & \multicolumn{4}{c}{Outlier Perturbation (ROC)} & \multirow{2}{*}{Max $\sigma$}\\
\cmidrule(lr){2-5} 
{} &         $5\%$ &     $10\%$ &     $20\%$ &     $40\%$ & {} \\
\midrule

Isolation Forest \cite{isolation_forest} &      $0.962$ &  $0.964$ &  $0.962$ &  $0.961$ &  $\pm0.02$ \\
INNE  \cite{algo-inne}           &      $0.972$ &  $0.975$ &  $\boldsymbol{0.975}^{+}$ &  $\boldsymbol{0.970}^{+}$ &  $\pm0.01$ \\
LOF   \cite{local_outlier_factor}           &      $0.970$ &  $0.972$ &  $0.970$ &  $\boldsymbol{0.970}^{+}$ &  $\pm0.01$ \\
AutoEncoder  \cite{Outlier_Analysis_aggarwal}    &      $0.971$ &  $0.970$ &  $0.968$ &  $0.965$ &  $\pm0.02$ \\
LUNAR   \cite{algo-lunar}         &      $0.972$ &  $0.969$ &  $0.968$ &  $0.966$ &  $\pm0.02$ \\
KNN Distance  \cite{knn-distance-algo}   &      $\boldsymbol{0.977}^{*}$ &  $\boldsymbol{0.978}^{*}$ &  $\boldsymbol{0.978}^{*}$ &  $\boldsymbol{0.977}^{*}$ &  $\pm0.01$ \\
GOAD  \cite{goad}           &      $0.969$ &  $0.970$ &  $0.969$ &  $0.969$ &  $\pm0.01$ \\
Deep SVDD  \cite{2018_deepsvdd}      &      $0.958$ &  $0.963$ &  $0.959$ &  $0.961$ &  $\pm0.02$ \\
RCA   \cite{rca}           &      $0.769$ &  $0.756$ &  $0.763$ &  $0.759$ &  $\pm0.11$ \\
NeuTraL  \cite{2021_neural_transformation}        &      $0.967$ &  $0.970$ &  $0.968$ &  $0.964$ &  $\pm0.02$ \\
RANSAC-NN (Ours)       &      $\boldsymbol{0.976}^{+}$ &  $\boldsymbol{0.976}^{+}$ &  $0.971$ &  $0.935$ &  $\pm0.03$ \\

\bottomrule
\end{tabular}
\end{adjustbox}

    \caption{\textbf{Outlier Detection with Clean Training (MobileNet-v3).}}
    \label{suppl:clean_train_mobilenet}
\end{table}

\begin{table}[h]
    \centering
    \begin{adjustbox}{width=0.47\textwidth}
    \begin{tabular}{lllllc}
\toprule
\multirow{2}{*}{Algorithm} & \multicolumn{4}{c}{Outlier Perturbation (ROC)} & \multirow{2}{*}{Max $\sigma$}\\
\cmidrule(lr){2-5} 
{} &         $5\%$ &     $10\%$ &     $20\%$ &     $40\%$ & {} \\
\midrule

Isolation Forest \cite{isolation_forest} &      $0.961$ &  $0.962$ &  $0.959$ &  $0.957$ &  $\pm0.02$ \\
INNE  \cite{algo-inne}           &      $0.974$ &  $0.972$ &  $0.971$ &  $0.968$ &  $\pm0.01$ \\
LOF  \cite{local_outlier_factor}            &      $0.966$ &  $0.963$ &  $0.965$ &  $0.963$ &  $\pm0.02$ \\
AutoEncoder \cite{Outlier_Analysis_aggarwal}     &      $0.974$ &  $0.974$ &  $0.972$ &  $\boldsymbol{0.970}^{+}$ &  $\pm0.02$ \\
LUNAR \cite{algo-lunar}           &      $0.973$ &  $0.973$ &  $\boldsymbol{0.974}^{+}$ &  $\boldsymbol{0.970}^{+}$ &  $\pm0.02$ \\
KNN Distance \cite{knn-distance-algo}    &      $\boldsymbol{0.980}^{+}$ &  $\boldsymbol{0.980}^{*}$ &  $\boldsymbol{0.979}^{*}$ &  $\boldsymbol{0.979}^{*}$ &  $\pm0.02$ \\
GOAD  \cite{goad}           &      $0.962$ &  $0.961$ &  $0.965$ &  $0.963$ &  $\pm0.02$ \\
Deep SVDD  \cite{2018_deepsvdd}      &      $0.923$ &  $0.925$ &  $0.931$ &  $0.929$ &  $\pm0.04$ \\
RCA  \cite{rca}            &      $0.831$ &  $0.832$ &  $0.809$ &  $0.792$ &  $\pm0.11$ \\
NeuTraL \cite{2021_neural_transformation}         &      $0.937$ &  $0.937$ &  $0.929$ &  $0.915$ &  $\pm0.06$ \\
RANSAC-NN (Ours)       &      $\boldsymbol{0.981}^{*}$ &  $\boldsymbol{0.976}^{+}$ &  $0.970$ &  $0.946$ &  $\pm0.03$ \\

\bottomrule
\end{tabular}
\end{adjustbox}

    \caption{\textbf{Outlier Detection with Clean Training (Fast-ViT).}}
    \label{suppl:clean_train_swin}
\end{table}

\subsection{Influence of Contaminated Training}
\begin{figure}[h]
    \centering
    \includegraphics[width=0.47\textwidth]{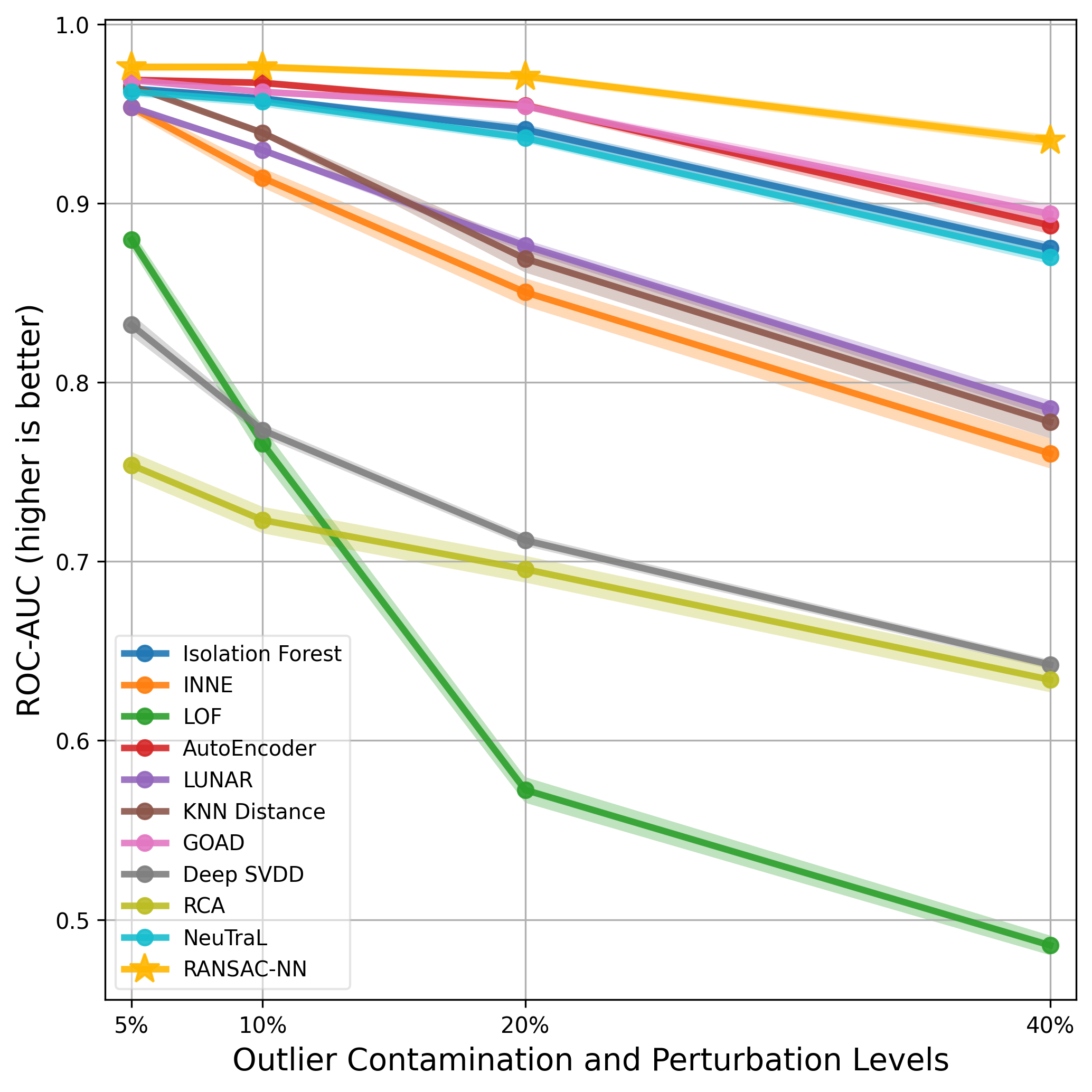}
    \caption{\textbf{Influence of Contaminated Training (MobileNet).}}
    \label{suppl:contam_train_mobilenet}
\end{figure}

\begin{figure}[h]
    \centering
    \includegraphics[width=0.47\textwidth]{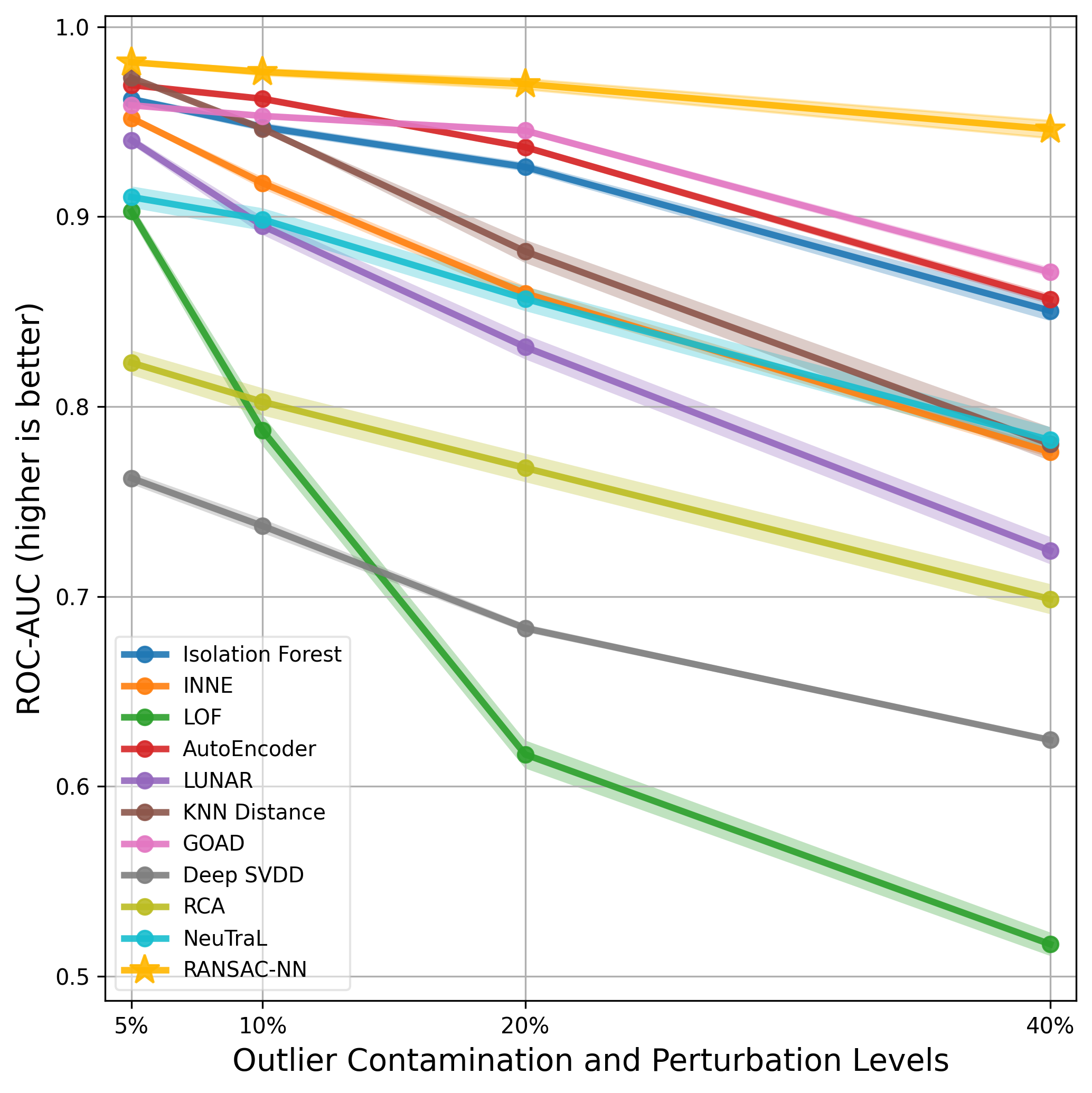}
    \caption{\textbf{Influence of Contaminated Training (Fast-ViT).}}
    \label{suppl:contam_train_fastvit}
\end{figure}

The same experiments from Section \ref{exp:impact_of_contaminated_training} are repeated below. As shown in Figures \ref{suppl:contam_train_mobilenet} and \ref{suppl:contam_train_fastvit}, the performance drop due to contaminated training can be observed regardless of the feature extractor choice.

\subsection{Outlier Filtering with RANSAC-NN}
\label{suppl:outlier_filtering}

\begin{table*}[h]
    \centering
    \begin{adjustbox}{width=\textwidth}
    \begin{tabular}{lllllll}
\toprule
\multirow{2}{*}{Algorithm} & \multicolumn{2}{c}{10\% Contamination} & \multicolumn{2}{c}{20\% Contamination} & \multicolumn{2}{c}{40\% Contamination}\\
\cmidrule(lr){2-3} \cmidrule(lr){4-5} \cmidrule(lr){6-7}
{} & Take Top-50\% & Take Top-90\% & Take Top-50\% & Take Top-80\% & Take Top-50\% & Take Top-60\%\\

\midrule
Isolation Forest & $0.937\:(-1.81\%)$ & $\boldsymbol{0.963\:(+0.79\%)}$ & $0.945\:(+0.01\%)$ & $\boldsymbol{0.958\:(+1.38\%)}$ & $\boldsymbol{0.944\:(+5.69\%)}$ & $0.930\:(+4.32\%)$ \\ 
INNE & $0.914\:(-0.83\%)$ & $\boldsymbol{0.962\:(+3.97\%)}$ & $0.929\:(+7.16\%)$ & $\boldsymbol{0.940\:(+8.35\%)}$ & $\boldsymbol{0.913\:(+14.33\%)}$ & $0.911\:(+14.15\%)$ \\ 
LOF & $0.935\:(+16.28\%)$ & $\boldsymbol{0.943\:(+17.05\%)}$ & $\boldsymbol{0.946\:(+37.24\%)}$ & $0.906\:(+33.29\%)$ & $\boldsymbol{0.911\:(+42.61\%)}$ & $0.884\:(+39.93\%)$ \\ 
AutoEncoder & $0.912\:(-5.61\%)$ & $\boldsymbol{0.970\:(+0.12\%)}$ & $0.923\:(-3.57\%)$ & $\boldsymbol{0.966\:(+0.73\%)}$ & $0.930\:(+3.97\%)$ & $\boldsymbol{0.935\:(+4.46\%)}$ \\ 
LUNAR & $0.900\:(-2.85\%)$ & $\boldsymbol{0.960\:(+3.16\%)}$ & $0.924\:(+4.10\%)$ & $\boldsymbol{0.952\:(+6.89\%)}$ & $\boldsymbol{0.922\:(+13.73\%)}$ & $0.922\:(+13.65\%)$ \\ 
KNN Distance & $\boldsymbol{0.975\:(+3.74\%)}$ & $0.968\:(+3.03\%)$ & $\boldsymbol{0.975\:(+10.35\%)}$ & $0.958\:(+8.65\%)$ & $\boldsymbol{0.942\:(+16.68\%)}$ & $0.940\:(+16.48\%)$ \\ 
GOAD & $\boldsymbol{0.970\:(+0.61\%)}$ & $0.969\:(+0.55\%)$ & $\boldsymbol{0.966\:(+0.94\%)}$ & $0.964\:(+0.76\%)$ & $\boldsymbol{0.958\:(+6.08\%)}$ & $0.950\:(+5.25\%)$ \\ 
Deep SVDD & $\boldsymbol{0.971\:(+20.00\%)}$ & $0.913\:(+14.22\%)$ & $\boldsymbol{0.966\:(+25.79\%)}$ & $0.894\:(+18.56\%)$ & $\boldsymbol{0.927\:(+28.41\%)}$ & $0.896\:(+25.33\%)$ \\ 
RCA & $\boldsymbol{0.786\:(+6.28\%)}$ & $0.746\:(+2.27\%)$ & $\boldsymbol{0.792\:(+8.69\%)}$ & $0.768\:(+6.36\%)$ & $\boldsymbol{0.774\:(+13.28\%)}$ & $0.754\:(+11.34\%)$ \\ 
NeuTraL & $\boldsymbol{0.962\:(+0.59\%)}$ & $0.961\:(+0.45\%)$ & $\boldsymbol{0.967\:(+2.81\%)}$ & $0.956\:(+1.76\%)$ & $\boldsymbol{0.953\:(+8.22\%)}$ & $0.942\:(+7.13\%)$ \\
\bottomrule
\end{tabular}
\end{adjustbox}
    \caption{\textbf{Performance Improvements from Outlier Filtering using RANSAC-NN (MobileNet-v3).}}
    \label{suppl:outlier_filtering_results_mbnet}
\end{table*}

\begin{table*}[h!]
    \centering
    \begin{adjustbox}{width=\textwidth}
    \begin{tabular}{lllllll}
\toprule
\multirow{2}{*}{Algorithm} & \multicolumn{2}{c}{10\% Contamination} & \multicolumn{2}{c}{20\% Contamination} & \multicolumn{2}{c}{40\% Contamination}\\
\cmidrule(lr){2-3} \cmidrule(lr){4-5} \cmidrule(lr){6-7}
{} & Take Top-50\% & Take Top-90\% & Take Top-50\% & Take Top-80\% & Take Top-50\% & Take Top-60\%\\

\midrule
Isolation Forest & $0.926\:(-2.29\%)$ & $\boldsymbol{0.954\:(+0.49\%)}$ & $0.929\:(+0.50\%)$ & $\boldsymbol{0.947\:(+2.27\%)}$ & $0.893\:(+4.25\%)$ & $\boldsymbol{0.896\:(+4.58\%)}$ \\ 
INNE & $0.916\:(-0.70\%)$ & $\boldsymbol{0.952\:(+2.86\%)}$ & $0.924\:(+6.10\%)$ & $\boldsymbol{0.926\:(+6.27\%)}$ & $\boldsymbol{0.885\:(+11.31\%)}$ & $0.882\:(+11.04\%)$ \\ 
LOF & $\boldsymbol{0.941\:(+15.57\%)}$ & $0.940\:(+15.46\%)$ & $\boldsymbol{0.946\:(+32.63\%)}$ & $0.894\:(+27.40\%)$ & $\boldsymbol{0.878\:(+35.95\%)}$ & $0.860\:(+34.16\%)$ \\ 
AutoEncoder & $0.918\:(-4.15\%)$ & $\boldsymbol{0.967\:(+0.82\%)}$ & $0.927\:(-1.10\%)$ & $\boldsymbol{0.959\:(+2.07\%)}$ & $0.905\:(+4.77\%)$ & $\boldsymbol{0.917\:(+5.98\%)}$ \\ 
LUNAR & $0.905\:(+1.22\%)$ & $\boldsymbol{0.954\:(+6.09\%)}$ & $0.920\:(+8.54\%)$ & $\boldsymbol{0.934\:(+9.93\%)}$ & $\boldsymbol{0.898\:(+16.74\%)}$ & $0.884\:(+15.32\%)$ \\ 
KNN Distance & $\boldsymbol{0.977\:(+2.94\%)}$ & $0.971\:(+2.37\%)$ & $\boldsymbol{0.971\:(+8.79\%)}$ & $0.952\:(+6.88\%)$ & $\boldsymbol{0.916\:(+13.32\%)}$ & $0.915\:(+13.26\%)$ \\ 
GOAD & $\boldsymbol{0.960\:(+0.65\%)}$ & $0.959\:(+0.53\%)$ & $\boldsymbol{0.964\:(+1.85\%)}$ & $0.959\:(+1.32\%)$ & $\boldsymbol{0.918\:(+4.57\%)}$ & $0.918\:(+4.57\%)$ \\ 
Deep SVDD & $\boldsymbol{0.953\:(+21.60\%)}$ & $0.857\:(+11.96\%)$ & $\boldsymbol{0.944\:(+26.04\%)}$ & $0.859\:(+17.55\%)$ & $\boldsymbol{0.876\:(+25.15\%)}$ & $0.839\:(+21.42\%)$ \\ 
RCA & $0.792\:(-1.01\%)$ & $\boldsymbol{0.821\:(+1.85\%)}$ & $0.781\:(+1.30\%)$ & $\boldsymbol{0.792\:(+2.45\%)}$ & $0.760\:(+6.11\%)$ & $\boldsymbol{0.763\:(+6.40\%)}$ \\ 
NeuTraL & $\boldsymbol{0.929\:(+3.04\%)}$ & $0.924\:(+2.58\%)$ & $\boldsymbol{0.920\:(+6.28\%)}$ & $0.910\:(+5.37\%)$ & $\boldsymbol{0.877\:(+9.49\%)}$ & $0.874\:(+9.18\%)$ \\ 
\bottomrule
\end{tabular}
\end{adjustbox}
    \caption{\textbf{Performance Improvements from Outlier Filtering using RANSAC-NN (Fast-ViT).}}
    \label{suppl:outlier_filtering_results_fastvit}
\end{table*}

\begin{figure*}[h!]
    \centering
    \includegraphics[width=0.98\textwidth]{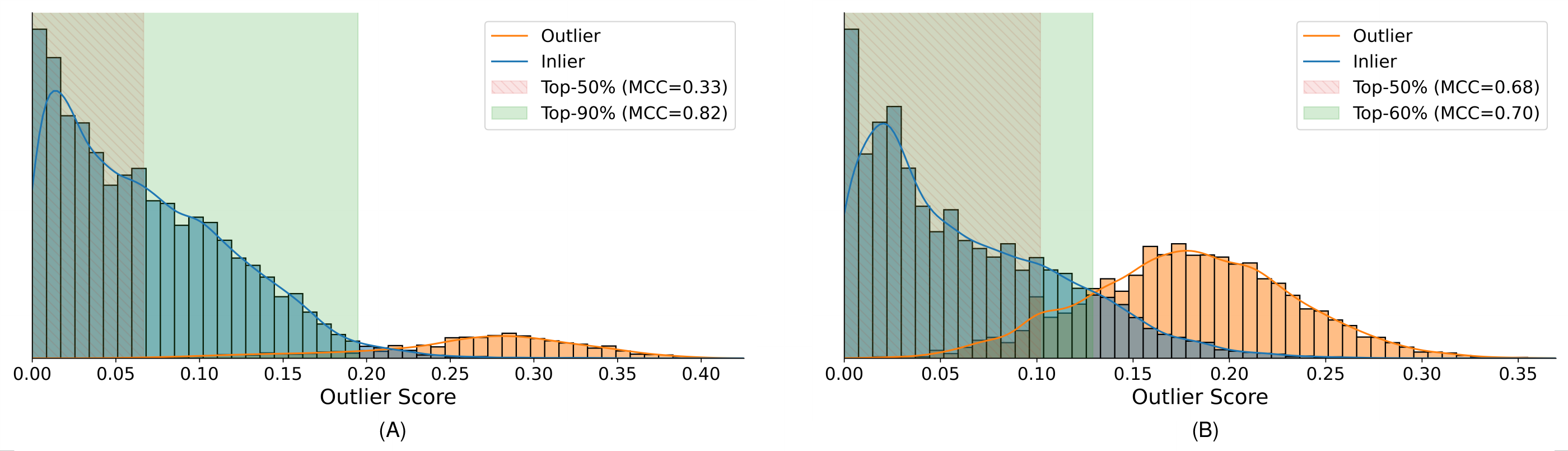}
    \caption{\textbf{Outlier Score Distributions.} Shown above are the outlier score distributions of a contaminated ImageNet21K dataset predicted by RANSAC-NN. Plots A \textit{(left)} and B (\textit{right}) illustrate the distributions under $10\%$ and $40\%$ contamination levels.}
    \label{fig:outlier_filtering_dist_comb}
\end{figure*}

The same experiments from Section \ref{exp:outlier_filtering} are repeated in the following. For each algorithm, we train a model on a set of images that had been outlier-filtered by RANSAC-NN prior to model training (see Figures \ref{fig:outlier_filtering_dist} and \ref{fig:outlier_filtering_dist_comb} for example illustration). The resulting model performance are listed in Tables \ref{suppl:outlier_filtering_results_mbnet} and \ref{suppl:outlier_filtering_results_fastvit} for both feature extractors. 

We can notice that when the outlier contamination level is severe, almost all algorithms benefit from outlier-filtering with a smaller but higher quality training dataset. This demonstrates the importance of training on a clean inlier set. Furthermore, these observations are consistent across different feature extractor choices.

\section{Run-Time Performance}
\begin{figure*}[t]
    \centering
    \includegraphics[width=0.74\textwidth]{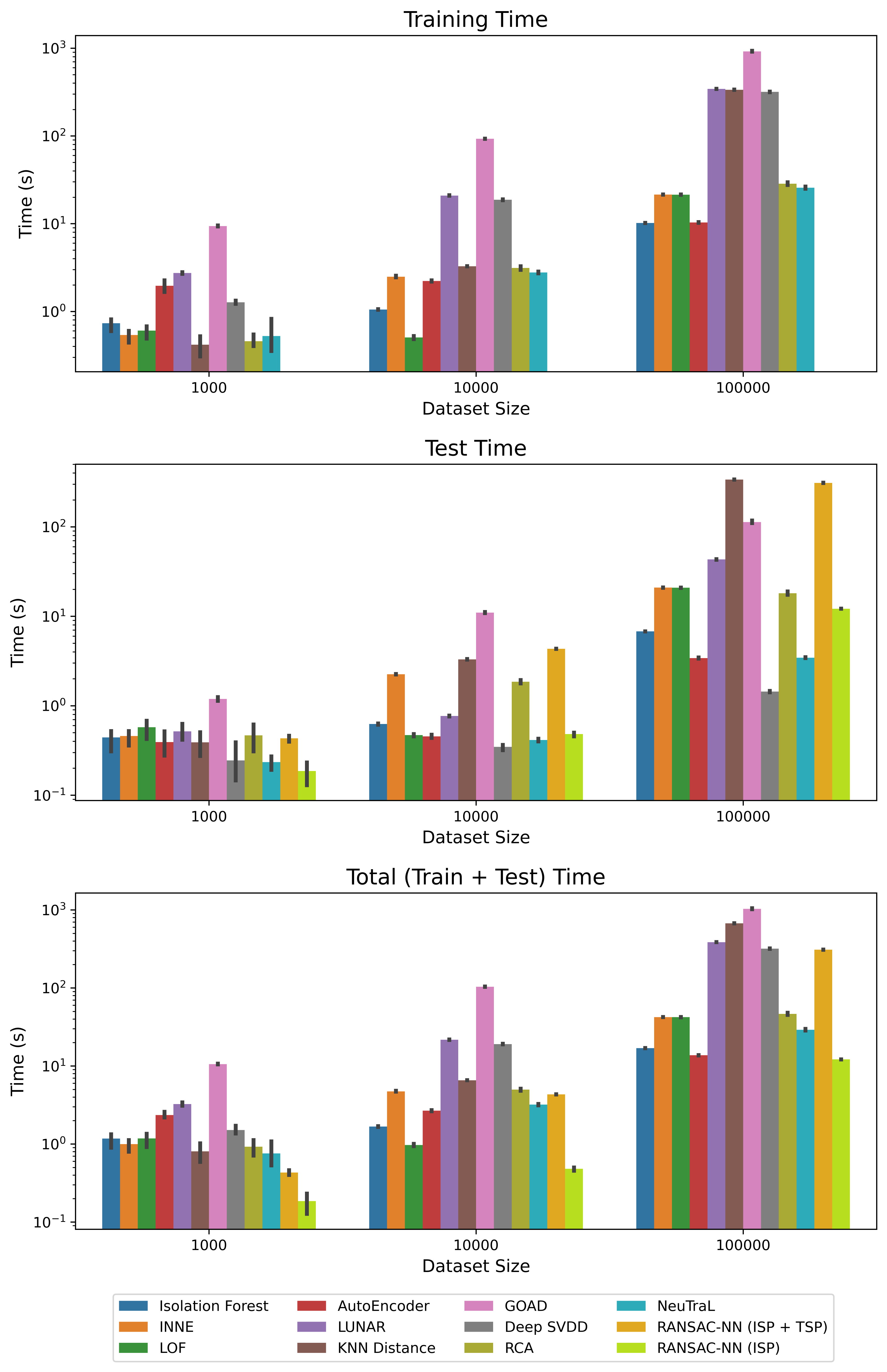}
    \caption{\textbf{Run-time Performance of OD Algorithms.} Shown above are the training, prediction, and total run-time of the OD algorithms from Section \ref{exp:experiment_setup}. Since RANSAC-NN is a unsupervised algorithm that is applied dynamically during test time, all of it computation takes place during test time. When considering the total computation time, RANSAC-NN maintains a moderate run-time in comparison of other algorithms. The run-time can be further reduced if only ISP is applied, but there would be a potential drop in prediction performance (see Section \ref{exp:ablation_studies} for details).}
    \label{suppl:runtime-exp}
\end{figure*}

We provide a comparison of the run-time of each of the algorithms from Section \ref{exp:experiment_setup} using the PyOD \cite{zhao2019pyod} and DeepOD \cite{deepod} libraries. Nearest-neighbors in RANSAC-NN was implemented using FAISS \cite{faiss}. Experiments were carried out on an Intel Core i7-8700K @ 3.7GHz CPU. We consider a feature dimension of $256$, and we evaluate the algorithms on datasets containing $1000$, $10000$, and $100000$ samples. The average run-time from 4 repeated experiments are shown in Figure \ref{suppl:runtime-exp}.  

\begin{figure*}[t!]
    \centering
    \includegraphics[width=\textwidth]{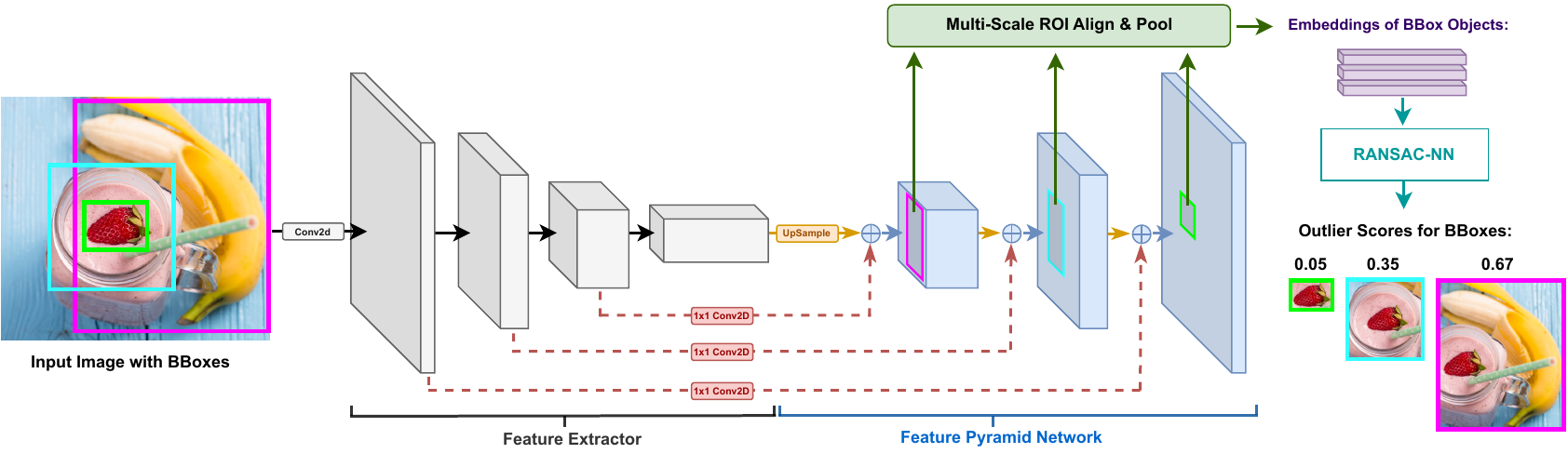}
    \caption{\textbf{Mislabeled Bounding Box Detection with RANSAC-NN.} Shown above is an illustration of our bounding box mislabeled detection algorithm with RANSAC-NN. Embeddings for bounding boxes are extracted by passing each image through a feature extractor followed by a Feature Pyramid Network as in \cite{fasterrcnn}. The Multi-Scale ROI Align and pooling operation extracts a fixed size embedding for each bounding box depending on its size. The embeddings of bounding boxes belonging to each category are then passed through RANSAC-NN for scoring. Bounding boxes with high outlier scores have a high likelihood of being mislabeled.}
    \label{suppl:bbox_mislabeled_detection}
\end{figure*}


\section{Applications In Image Mislabeled Detection}

Since RANSAC-NN operates as a one-class classifier, its applications in image mislabeled detection may also be explored. In the case of \textbf{image-level} labels, one may apply RANSAC-NN on the set of images belonging to every class. Images that are mislabeled may be detected with potentially high outlier scores. In the case of \textbf{bounding-box} labels, one may apply RANSAC-NN on the crops of bounding boxes from every object class. Mislabeled bounding boxes may be correlated with abnormally high outlier scores. A potential approach using the Faster-RCNN \cite{fasterrcnn} architecture is illustrated in Figure \ref{suppl:bbox_mislabeled_detection}.

\end{document}